\documentclass{article}

 \usepackage[main, final]{neurips_2026}

\usepackage[utf8]{inputenc} 
\usepackage[T1]{fontenc}    
\usepackage{hyperref}       
\usepackage{url}            
\usepackage{booktabs}       
\usepackage{amsfonts}       
\usepackage{nicefrac}       
\usepackage{microtype}      
\usepackage{xcolor}         
\usepackage{graphicx}
\usepackage{subfigure}
\usepackage{hyperref}
\usepackage{url}
\usepackage{lipsum}
\usepackage{listings}
\usepackage{enumitem}
\usepackage{colortbl}
\usepackage{pifont}
\usepackage{subcaption}
\usepackage{wrapfig}
\usepackage[most]{tcolorbox}
\usepackage{makecell}
\usepackage[dvipsnames]{xcolor}
\usepackage{cleveref} 
\usepackage{multicol}
\usepackage{multirow}
\usepackage{kotex}
\usepackage{threeparttable}
\usepackage{tcolorbox}
\tcbuselibrary{breakable}

\newcommand{\proposeddata}{\textsf{PerMem-Bench}}
\newcommand{\proposed}{\textit{Personalize-then-Store}}

\newtcolorbox{promptbox}[1][]{
  breakable,
  colback=gray!8,
  colframe=gray!50,
  fonttitle=\bfseries\small,
  title=#1,
  left=6pt, right=6pt, top=4pt, bottom=4pt,
  boxrule=0.5pt,
}

\title{\proposed: Benchmarking and Learning Personalized Memory for Long-horizon Agents}

\author{
\textbf{Yeonjun In}, 
\textbf{Wonjoong Kim}, 
\textbf{Sangwu Park}, 
\textbf{Kanghoon Yoon}, 
\textbf{Chanyoung Park}\thanks{Corresponding Author} \\
KAIST \\
\texttt{\{yeonjun.in, wjkim, sangwu.park, ykhoon08, cy.park\}@kaist.ac.kr} \\
}

\begin{document}

\maketitle

\begin{abstract}
Existing large language model (LLM)-based memory systems apply universal, static policies that overlook a fundamental reality: the contexts that are worth storing in memory are different across users. This misalignment wastes limited memory budget on transient interactions while failing to preserve critical context for long-horizon tasks. To address this gap, we investigate an underexplored question: can LLM-based memory systems learn personalized memory policies? We introduce \proposeddata, the first benchmark that jointly models multi-domain interaction histories and cross-user heterogeneity in agent-use patterns to evaluate personalized memory systems. We further present the first empirical study of memory personalization, proposing \textit{session-level storage gating} — a lightweight framework that selectively bypasses memory operations for transient sessions. Our study confirms that personalization yields substantial retention gains under perfect gating, yet reveals that accurate gating remains an open and critical challenge. 
Our benchmark and source code are available at \href{https://github.com/yeonjun-in/PerMemBench}{\textcolor{magenta}{https://github.com/yeonjun-in/PerMemBench.}} 

\end{abstract}

\section{Introduction}

The proliferation of LLM agents has attracted diverse users tasking agents with both transient and long-horizon interactions across various domains. Unlike transient tasks, successful long-horizon interactions require agents to preserve and manage crucial context from past interactions. Since LLMs inherently lack the capacity to memorize prior context, memory systems have emerged as a cornerstone for sustaining effective and coherent long-horizon agent-user dialogues. \cite{chhikara2025mem0, zhou2025mem1, yan2025memoryr1, xu2025amem, yang2026plugmem, packer2023memgpt, wang2025memalpha}.

Early memory systems relied on storing exhaustive raw dialogue histories within a memory bank or context window. However, this naive approach is impractical for real-world deployment, as it necessitates an infinite memory budget and introduces substantial irrelevant noise. Subsequent research has shifted focus toward deliberately extracting critical information to operate within a fixed budget \cite{hu2025memorysurvey}. 
Specifically, LLM agents are trained to identify "worth-storing" contexts—i.e., information whose preservation is expected to benefit future interactions, such as user preferences or specific events—and to update or delete existing memories via in-context learning or post-training \cite{chhikara2025mem0, tan2025prospect, zhou2025mem1, yan2025memoryr1, xu2025amem}.
These trained policies apply a universal, one-size-fits-all memory system to all users, regardless of individual differences.

\begin{figure*}[h] 
    \centering
    \includegraphics[width=.95\linewidth]{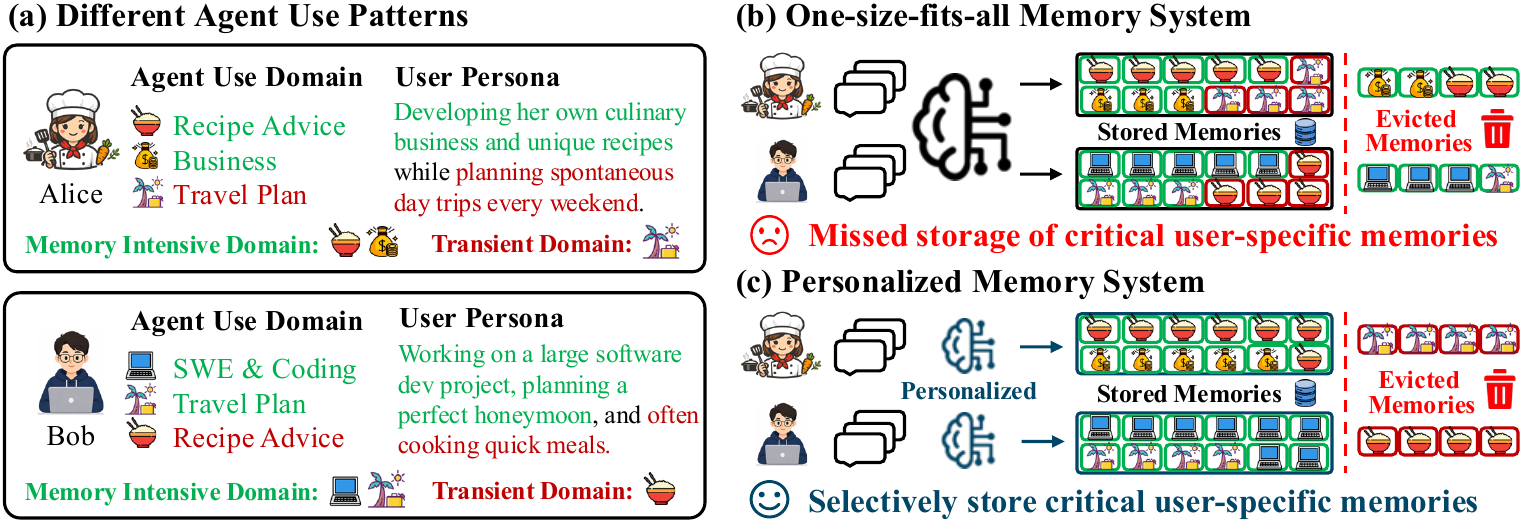}
    \vspace{-1ex}
    \caption{Motivating examples of personalized memory system. (a) Users exhibit distinct agent use patterns. (b) One-size-fits-all memory systems fail to personalize these user-specific needs, leading to the eviction of essential contexts. (c) An ideal personalized memory policy selectively preserves essential contexts tailored to each user's use pattern.}
    \vspace{-4ex}
    \label{fig:motivation}
\end{figure*}

However, this paradigm overlooks a fundamental question: \textbf{Are the contexts that are worth storing in memory the same for all users?} As illustrated in \Cref{fig:motivation}(a), users exhibit heterogeneous agent use patterns across various domains. For Alice, `Recipe Advice' is a long-horizon project requiring consistent context preservation, whereas `Travel Plan' involves only spontaneous, transient inquiries. Conversely, Bob regards `Travel Plan' as a memory-intensive long-horizon task for honeymoon planning, while his `Recipe Advice' usage is strictly transient. Consequently, the information within a `Travel Plan' interaction constitutes a "worth-storing" context for Bob but not for Alice—and the inverse holds true for `Recipe Advice'.

We observe that existing memory systems fail to account for these heterogeneous \textbf{user-specific patterns}, instead managing memory based on universal criteria. This leads to a critical \textbf{misallocation of resources}: the system wastes limited memory budget on transient interactions while failing to preserve essential context for vital long-horizon tasks (see \Cref{fig:motivation}(b)).
To address this, we argue that an ideal memory system should be \textbf{personalized}, where the system should infer the user-specific "worth-storing" contexts then selectively store them—bypassing unnecessary storage for transient interactions while prioritizing those requiring long-horizon context accumulation (see \Cref{fig:motivation}(c)). 

Regarding this observation, we raise an important yet underexplored research question: \textbf{Can LLM-based memory systems infer the user-specific "worth-storing" contexts and learn personalized policy? }However, there is no benchmark dataset featuring long-horizon dialogues that capture the heterogeneous and personalized usage patterns observed across diverse users and domains. This absence precludes a rigorous evaluation of a memory system's capacity for fine-grained personalization. 

To this end, we introduce \proposeddata, a novel benchmark for evaluating personalized memory systems, along with a fully automated data generation pipeline. The pipeline proceeds in three stages: (1) profiling user-specific agent use patterns for diverse personas, (2) constructing a {life skeleton} per user — a structured blueprint defining their long-horizon interaction trajectory — and (3) synthesizing realistic dialogue sessions via an LLM-based user simulator. By assigning a unique agent use profile to each persona, we instantiate user-specific ``worth-storing'' contexts, enabling rigorous evaluation of whether a memory system can accurately infer and preserve information tailored to each individual. The resulting dataset comprises multi-year interaction sessions for 20 users spanning diverse domains. Since the pipeline is fully automated and requires no manual intervention, it can be readily scaled to larger and more diverse user cohorts beyond the current set.

Building on \proposeddata, we investigate our research question through a systematic empirical study. We propose \textit{session-level storage gating}, a simple yet general personalization framework that identifies whether each session is long-horizon or transient and skips memory operations for the latter, and introduce multiple gating methods as baselines. Our experiments show 
that perfect gating yields substantial retention and downstream performance gains under a fixed budget, yet current 
baselines remain suboptimal in gating accuracy, achieving only incremental gains in 
practice. These results illuminate the difficulty of personalizing memory systems in the 
wild and provide concrete directions for future research.

\noindent Our contributions are as follows:
\begin{itemize}[leftmargin=0.5cm]
  \item We identify and formalize the critical need for personalized memory systems, 
  moving beyond the current ``one-size-fits-all'' paradigm.
  
  \item We present \proposeddata, the first benchmark that jointly models multi-domain interaction histories and cross-user heterogeneity in agent-use patterns to evaluate personalized memory systems.
  
  \item We introduce the first empirical study on memory personalization, proposing 
  \textit{session-level storage gating} as a novel personalization paradigm and establishing 
  simple baselines as a reference point for future work.
\end{itemize}

\section{Related Work}
\label{sec:related-work}

\textbf{Agent Memory Systems.} \@ 
AI agents increasingly rely on memory systems to support long-horizon tasks across diverse users. Recent research in this area can be broadly categorized into two directions. The first direction focuses on learning LLM-based memory policies, enabling them to selectively extract and store salient information from interactions \cite{chhikara2025mem0, zhou2025mem1, yan2025memoryr1, tan2025prospect, memobase}. A central challenge in this line of work is determining which information is worth storing for a user. The second direction focuses on structured memory representations, leveraging clustering, graph, and tree-based methods to model relationships among memory units and improve retrieval accuracy \cite{xu2025amem, yang2026plugmem, hu2025memorysurvey, rezazadeh2024memtree, chhikara2025mem0, hipporag2}.

Our work aligns with the first direction but distinguishes itself by moving beyond the uniform criteria of prior approaches. Rather than applying a universal standard for identifying information worth storing, we propose \textit{session-level storage gating} as a novel personalization paradigm that learns to identify each user's worth-storing sessions from their interaction history, and selectively bypasses memory operations for transient ones.

\textbf{Evaluation of Agent Memory Systems.} \@
Evaluation frameworks for agent memory are typically divided into experiential and factual memory: the former distills past interactions into skills and strategies for improved reasoning, while the latter focuses on preserving critical user-centric context over long-horizon interactions. This study focuses on the latter, specifically evaluating whether a memory system effectively stores "worth-storing" information tailored to a user. Existing benchmarks in this space adopt LLM-based user simulations to model realistic interactions and assess memory capabilities \cite{maharana2024locomo, kim2024dialsim, wu2024longmemeval, jiang2025personamem, chen2025halumem, jiayang2026amemgym}. 

However, we argue that these evaluations largely rely on unrealistic assumptions. First, most benchmarks impose a single-domain constraint. LoCoMo \cite{maharana2024locomo} and HalluMem \cite{chen2025halumem} focus exclusively on casual interactions in which users share personal events with agents, whereas real-world users engage with agents across multiple heterogeneous domains and goal-oriented scenarios. Second, they overlook behavioral heterogeneity across users. Benchmarks such as PersonaMem \cite{jiang2025personamem}, LongMemEval \cite{wu2024longmemeval}, and AmemGym \cite{jiayang2026amemgym} incorporate only \textbf{personal attributes}—such as demographics, traits, and preferences—as user profiles, while ignoring \textbf{behavioral attributes}, i.e., agent use patterns. As a result, these benchmarks implicitly assume all users exhibit homogeneous agent use pattern, failing to capture the meaningful differences that arise in real-world user–agent interactions.

To bridge these gaps, we introduce a new benchmark, \proposeddata, that captures these complex, real-world usage scenarios. Unlike prior work, \proposeddata~features multi-domain interaction histories and explicitly models \textit{behavioral heterogeneity}. This provides a rigorous environment for evaluating whether memory systems can be effectively personalized across diverse users with heterogeneous interaction patterns.

\section{Benchmark Construction: \proposeddata$^s$}
\vspace{-1ex}

This section details the construction of \proposeddata$^s$, a fully automated pipeline comprising three primary stages: (1) user-specific agent use profiling (\Cref{sec:agent-use-profiling}), (2) life skeleton and timeline construction (\Cref{sec:life-timeline-construction}), and (3) dialogue generation (\Cref{sec:dialogue-generation}).
\proposeddata$^s$ encompasses diverse agent use scenarios for 20 unique users. This sample size was strategically determined to balance the computational overhead of generation with the subsequent costs of memory system evaluation. While the current scale is optimized for efficiency, the inherent reliability of our automated process facilitates seamless scaling to larger cohorts, as discussed in \Cref{sec:meta-eval-and-dataset-quality}.  Importantly, \proposeddata~is constructed entirely from synthetic data, including synthetically generated personas and interaction histories. It contains no real-user interaction data or personally identifiable information.

\begin{figure*}[h] 
    \centering
    \vspace{-1ex}
    \includegraphics[width=.95\linewidth]{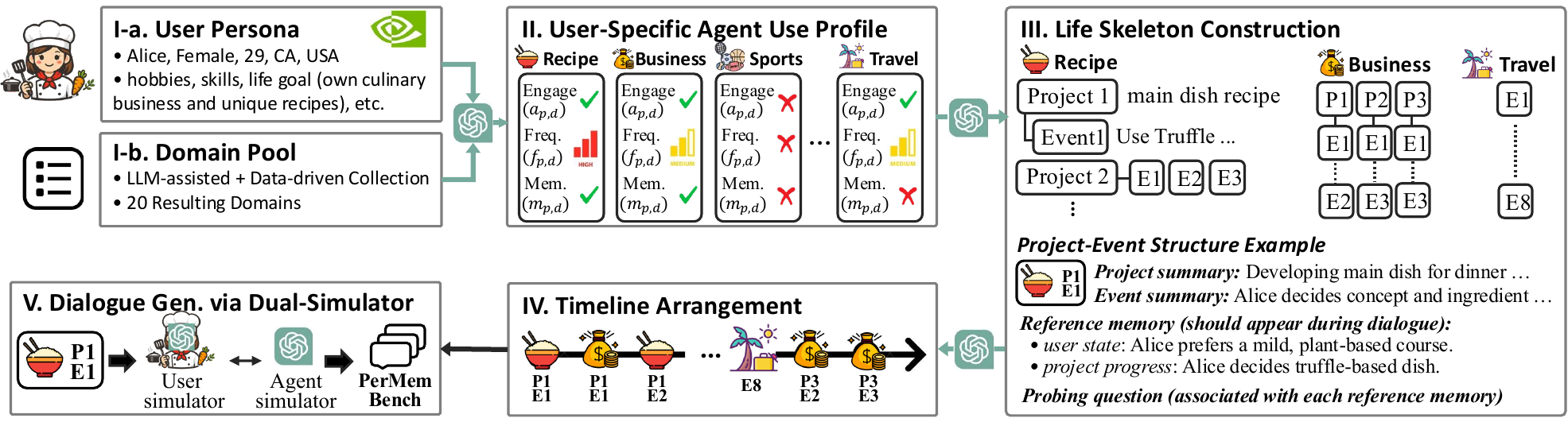}
    \vspace{-1ex}
    \caption{Overview of the construction pipeline for \proposeddata$^s$.}
    \vspace{-1ex}
    \label{fig:overview_benchmark}
\end{figure*}

\subsection{User-Specific Agent Use Profiling}
\label{sec:agent-use-profiling}

We define an agent use profile as the joint configuration of \textit{domain participation} and \textit{memory necessity} across domains. We posit these two dimensions offer a simple yet effective framework for capturing the diverse use patterns. For instance, Alice and Bob in \Cref{fig:motivation}(a) demonstrate divergent profiles under this framework. While we recognize more granular patterns exist, we adopt this simple setup as a foundational step toward establishing a baseline for personalized memory management.

\textbf{User Persona Collection (I-a of \Cref{fig:overview_benchmark}).} \@
To ensure real-world plausibility, we leverage the \textit{Nemotron-Persona-USA} dataset \cite{Nemotron-Personas-USA}. This collection provides high-fidelity personas with detailed attributes, including personal/professional backgrounds, personal preferences, allowing us to simulate a broad spectrum of user behaviors.

\textbf{Domain Pool Construction (I-b of \Cref{fig:overview_benchmark}).} \@
To ensure representative coverage of real-world usage, we employ a data-driven approach to construct a domain pool. First, we sample 1,000 personas and prompt \texttt{Claude Haiku 4.5} to generate potential usage scenarios without predefined constraints (see Appendix~\ref{sec:ap-domain-pool}). These candidates are then semantically clustered and assigned representative labels via human review. To align the pool with actual LLM trends, we cross-reference these clusters with industry reports \cite{chatterji2025people, openai2026chatgpt_usage_adoption_work}, pruning niche cases and supplementing broad-interest domains. This process results in a final taxonomy of 20 domains (see \Cref{tab:added_domains} in Appendix).

\textbf{User-Specific Profile Assignment (II of \Cref{fig:overview_benchmark}).} \@
From the collected persona set, we randomly sample 20 personas. For each persona $p \in \mathcal{P}$ and domain $d \in \mathcal{D}$, we employ \texttt{Claude-Haiku-4.5} to infer profiles based on the user’s lifestyle and objectives (see Appendix~\ref{sec:ap-profile-assign} for prompt details). This results in a triplet $\mathcal{M}_{p,d} = (a_{p,d}, f_{p,d}, m_{p,d})$ for every domain:

\begin{itemize}[leftmargin=0.5cm]
    \item \textbf{Domain Participation} ($a_{p,d} \in \{0, 1\}$): Whether the user with $p$ uses an agent in domain $d$.
    \item \textbf{Frequency} ($f_{p,d} \in \{\text{high}, \text{mid}, \text{low}\}$): How often the user interacts within this domain.
    \item \textbf{Memory Necessity} ($m_{p,d} \in \{0, 1\}$): Requirement for context preservation. Crucially, $m_{p,d}$ is determined by user-specific intent rather than inherent domain properties. 
\end{itemize}

We cross-verify the plausibility of the generated profiles using an ensemble of \texttt{GPT-5.1}, and \texttt{o3-mini}. Any domain is excluded from the user's profile if any model flag its metadata ($a_{p,d}$, $f_{p,d}$, or $m_{p,d}$) as implausible
We sample a set $\mathcal{S}p$ of $s$ domains from the persona’s active pool $\mathcal{D}_{act}^p = {d \mid a_{p,d}=1}$, ensuring a balanced distribution between domains with $m_{p,d}=1$ and $m_{p,d}=0$, thereby forming the final user-specific profile metadata.

\subsection{Life Skeleton and Timeline Construction (III and IV of \Cref{fig:overview_benchmark})}
\label{sec:life-timeline-construction}

Based on user-specific profiles, we utilize \texttt{gpt-5.4} to construct a life skeleton, a structured blueprint for simulating long-horizon user-agent interactions. 

For domains requiring memory ($m_{p,d}=1$), interactions are organized as a sequence of interconnected `projects'. Each project consists of multiple events, each corresponding to a single dialogue session. An event includes an interaction summary, reference memories, and probing questions. Reference memories represent "worth-storing" information, such as user states and project progress, and serve as the gold standard that the memory system is expected to capture. 
Each reference memory is paired with a probing question designed to be answerable only if the corresponding information has been stored, enabling the evaluation of downstream memory retrieval and utilization. For transient domains ($m_{p,d}=0$), interactions consist of independent events covering unrelated topics, without project-level dependencies and reference memories. The number of projects and events is determined by the frequency metadata ($f_{p,d}$).

Once the per-domain skeletons are established, an \texttt{gpt-5.4} arranges all events into a coherent, unified timeline. This integrated timeline provides the temporal and contextual structure needed to synthesize multi-turn dialogues that reflect a coherent and personalized long-horizon user experience. Please refer to Appendix~\ref{sec:ap-phase1-skeleton} and \ref{sec:ap-phase1-timeline} for detailed descriptions of the process.

\subsection{Dialogue Generation via Dual-Simulator (V of \Cref{fig:overview_benchmark})}
\label{sec:dialogue-generation}

Using the life skeleton and integrated timeline, we synthesize realistic interactions through a dual-simulator framework. The user simulator generates context-driven utterances by manifesting the attributes—such as user state and project progress—defined in each event. In contrast, the agent simulator operates without prior access to the skeleton, responding solely based on the user's input and its internal memory. This process yields a long-horizon dialogue corpus that reflects the diverse and personalized requirements of agent use. Detailed procedure is presented in Appendix~\ref{sec:ap-dialogue-generation}.

\section{Reflecting Shifts in Agent Use Profiles: \proposeddata$^d$}In real-world scenarios, user interests are often dynamic rather than static, evolving in response to significant life events such as career changes, new hobbies, or the conclusion of long-horizon projects. Such transitions inevitably lead to shifts in the user's agent use profiles. In this section, we describe the construction of \proposeddata$^d$, which simulates these profile shifts building upon the foundation of \proposeddata$^s$.

To model these transitions, we modify the user's predefined agent use profile by introducing additional domains from the previously unselected pool ($\mathcal{D}_{act}^p \setminus \mathcal{S}_p$), covering both memory-intensive ($m_{p,d}=1$) and transient ($m_{p,d}=0$) domains. Furthermore, we transition an existing domain in $\mathcal{S}_p$ from $m_{p,d}=1$ to $m_{p,d}=0$, reflecting the completion of a long-horizon project and its shift toward transactional interaction.

Based on this shifted profile, we leverage \texttt{gpt-5.4} to infer plausible life events that justify these transitions and construct a continued life skeleton following the methodology in \Cref{sec:life-timeline-construction}. The resulting post-shift skeleton is arranged into a timeline and seamlessly appended to the pre-shift sequence. Finally, we perform dialogue generation using the same dual-simulator framework as described in \Cref{sec:dialogue-generation}, yielding a continuous, long-horizon trajectory that reflects the user's evolving interests and agent use profiles. Please refer to Appendix~\ref{sec:ap-phase2-skeleton} for detailed descriptions of the process.

\section{Data Analysis and Meta Evaluation on \proposeddata}
\label{sec:meta-eval-and-dataset-quality}

\subsection{Data Analysis}

In this section, we provide an exploratory analysis of \proposeddata. \Cref{tab:dataset_statistics} summarizes the core statistics for both \proposeddata$^s$ (Static) and \proposeddata$^d$ (Dynamic). 

\begin{wrapfigure}{r}{0.55\textwidth} 
    \centering
    \vspace{-12pt}
    \begin{threeparttable}
        \captionof{table}{Statistics of \proposeddata$^s$ and \proposeddata$^d$. Min, Max, and Avg are computed across 20 users.}
        \label{tab:dataset_statistics}
        \scriptsize
        \setlength{\tabcolsep}{3pt}
        \begin{tabular}{lccc c ccc}
            \toprule
            & \multicolumn{3}{c}{\proposeddata$^s$} & & \multicolumn{3}{c}{\proposeddata$^d$} \\
            \cmidrule{2-4} \cmidrule{6-8}
            \textbf{Metric} & \textbf{Min} & \textbf{Max} & \textbf{Avg} & & \textbf{Min} & \textbf{Max} & \textbf{Avg} \\
            \midrule
            \# Sessions & 26 & 78 & 54 & & 62 & 148 & 104 \\
            Timeline (mo) & 15 & 20 & 17 & & 25 & 32 & 28 \\
            Total Tokens & 126K & 522K & 314K & & 340K & 1M & 634K \\
            Avg. Tokens / Sess. & 3.9K & 7.7K & 5.8K & & 3.6K & 8.1K & 6.1K \\
            \# Ref. Memories & 38 & 72 & 53 & & 78 & 146 & 97 \\
            \bottomrule
        \end{tabular}
    \end{threeparttable}
    \vspace{-10pt}
\end{wrapfigure}

Our simulation spans extensive timelines, covering up to 20 months in \proposeddata$^s$ and 32 months in \proposeddata$^d$, with up to 1M dialogue-history tokens per user. Individual sessions contain up to 8K tokens, largely driven by detailed agent utterances commonly observed in real-world applications. These dense long-context environments challenge memory systems to distinguish worth-storing information from noise. In total, \proposeddata\ includes up to 146 reference memories and probing questions per user and provides over 1,000 evaluation examples in \proposeddata$^s$ and nearly 2,000 in \proposeddata$^d$.

\begin{wrapfigure}{t}{0.55\linewidth}
    \centering
    \vspace{6ex}
    \includegraphics[width=1\linewidth]{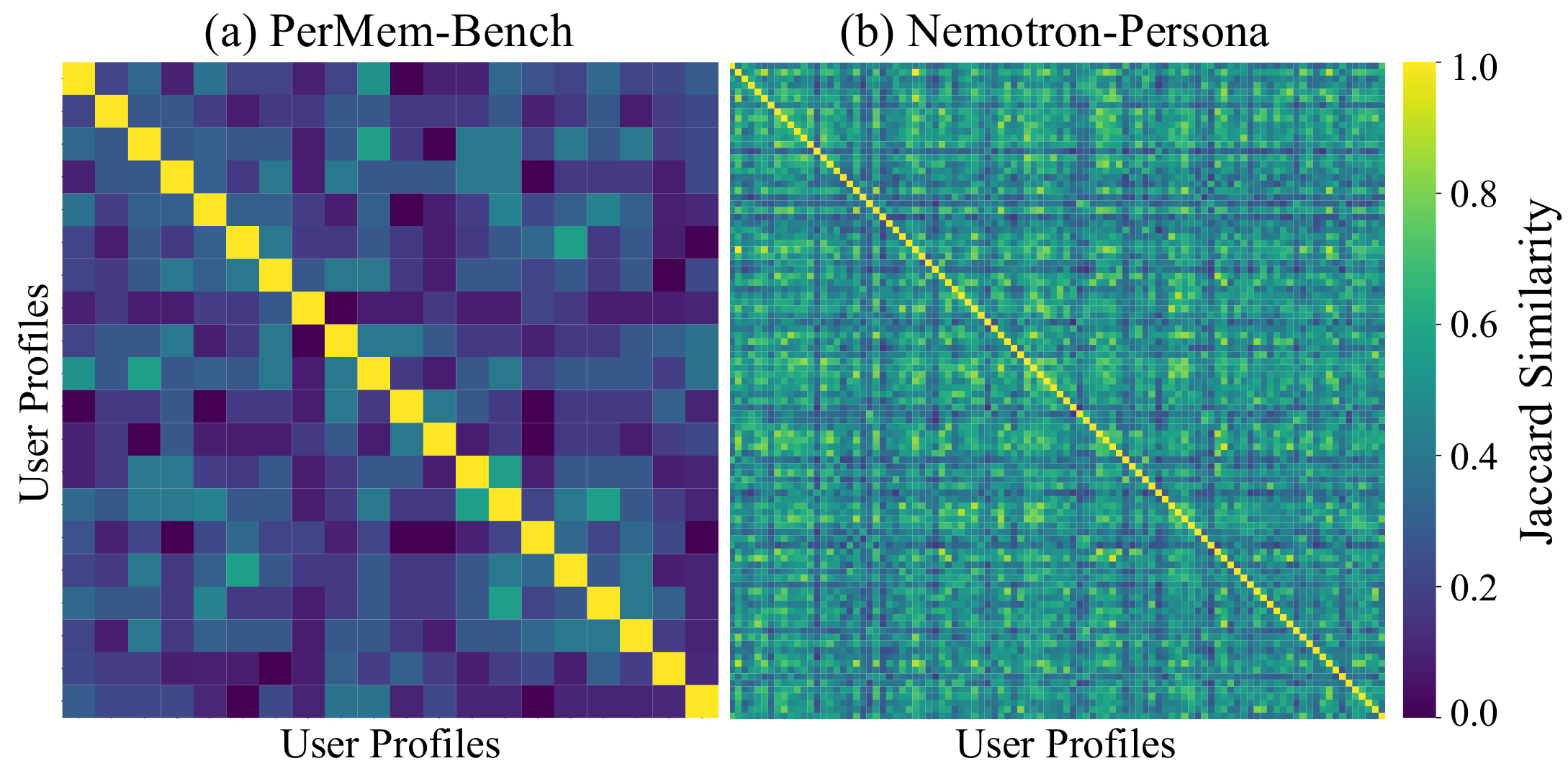}
    \caption{Similarity analysis on cross-user agent use profile. (a) Results on 20 users on \proposeddata. (b) Results on random 100 users from Nemotron-Persona. }
    \vspace{-8ex}
    \label{fig:dataset_stat}
\end{wrapfigure}

To ensure the diversity of the generated agent use profiles, which are defined by the combination of active domains and their respective memory necessity, we calculated the Jaccard Similarity between users, treating the domain-memory necessity pairs as features. A similarity value of 1 would indicate identical agent use patterns. As shown in \Cref{fig:dataset_stat}(a), the majority of pairs exhibit very low similarity, with no identical profiles existing in the set. To validate the scalability of this diversity, we sample 100 additional personas from the Nemotron-Persona-USA dataset and generate profiles using our pipeline. As illustrated in \Cref{fig:dataset_stat}(b), the results consistently demonstrate highly diverse use profiles. These findings confirm that \proposeddata\ effectively covers a broad spectrum of user behaviors in real-world agent application.

\subsection{Meta Evaluation}
To ensure the integrity of our data generation pipeline, we conduct 
a three-stage meta-evaluation. For each stage, we employ a panel of 
two evaluators—one human expert and one strong LLM judge 
(\texttt{Claude Opus 4.6})—and report the averaged quality score 
alongside inter-evaluator agreement measured by Gwet's 
AC1~\cite{gwet2001handbook}. 
Full details are provided in Appendix~\ref{sec:ap-meta-evaluation}.

\textbf{Stage 1: Profile Plausibility.}
We assess whether the generated agent use profiles are logically 
consistent with the assigned user personas, evaluating both relevance 
and realism. The panel achieves an average quality score of $99.5\%$ 
with an inter-evaluator agreement of $99.0\%$, indicating strong 
alignment between the generated profiles and the intended personas.

\textbf{Stage 2: Life Skeleton and Timeline Realism.}
We evaluate the coherence of project sequences and event timelines, 
verifying that reference memories are appropriate for the user persona 
and that temporal progressions are realistic. Both evaluators reach 
perfect agreement, with a quality score and AC1 of $100\%$.

\textbf{Stage 3: Dialogue Quality.}
We randomly sample 100 dialogue sessions and evaluate them along two 
dimensions: consistency with the life skeleton and seamless integration 
of reference memories. The panel achieves a quality score of $98.4\%$ 
with an inter-evaluator agreement of $96.9\%$, confirming that the 
synthesized dialogues are faithful to the predefined life trajectories.

Collectively, these results validate the reliability of our fully 
automated generation pipeline. Since the pipeline requires no manual 
intervention, \proposeddata~can be readily scaled to larger and more 
diverse user cohorts beyond the current 20-user set. 
The human evaluation for domain review and meta-evaluation was conducted internally by an author who is a Ph.D. student specializing in AI. No external evaluators were recruited, and no financial compensation was provided.

\section{Evaluation Protocol of \proposeddata}
\label{sec:evaluation-protocol}


An effective memory system must not only accurately extract, store, and persistently retain user-specific ``worth-storing'' contexts, but also retrieve and use them when they are needed for downstream tasks. Accordingly, \proposeddata\ evaluates memory systems along two complementary dimensions: whether they preserve tailored reference memories throughout their intended lifespans, and whether these memories remain accessible and useful for various downstream tasks. 

\subsection{Memory Storage Evaluation: Memory Retention Rate}

We leverage the Memory Retention Rate (RR), a 
metric that measures how consistently a reference memory unit remains 
in the memory bank throughout its required lifespan. We categorize 
lifespans based on the nature of the information: user-centric states 
(e.g., stable preferences or permanent attributes) must be retained 
until a relevant update occurs or the timeline concludes, whereas 
project-specific progress (e.g., decisions or milestones) must be 
retained at least until the corresponding project concludes.

Formally, let $\mathcal{R}$ be the set of reference memory units. 
For each $r \in \mathcal{R}$, we define $t_\text{start}(r)$ as the 
session at which the information first appears in the dialogue, making 
it eligible for storage, and $T_\text{target}(r)$ as its target 
retention horizon determined by the information type above. 
The Memory Retention Rate is:

\vspace{-3ex}
\begin{equation}
\small
RR = \frac{\sum_{r \in \mathcal{R}} \sum_{t=t_\text{start}(r)}^{T_\text{target}(r)} \mathbb{I}(r \in \mathcal{M}_t)}{\sum_{r \in \mathcal{R}} \left( T_\text{target}(r) - t_\text{start}(r) + 1 \right)}.
\label{eq:rr}
\end{equation}
\vspace{-2ex}

where $\mathcal{M}_t$ denotes the memory bank state at session $t$, 
and $\mathbb{I}(\cdot)$ is an indicator that equals 1 if $r$ is 
present in $\mathcal{M}_t$ and 0 otherwise.

\textbf{Practical Implementation.} \@
To determine $\mathbb{I}(r \in \mathcal{M}_t)$, we adopt an 
LLM-as-a-judge framework (\texttt{gpt-5-nano} is used) that verifies whether $r$ is preserved in $\mathcal{M}_t$. Rather than exhaustively checking all entries, the judge considers only the top-10 semantically similar entries retrieved from 
$\mathcal{M}_t$ using $r$ as the query, and performs a binary verdict. To validate the reliability of our LLM judge (gpt-5-nano), we randomly sampled 100 (reference memory, memory bank state) instances and had a human annotator independently judge whether each reference memory was present, following the same criteria as the judge prompt. Comparing human judgments against the LLM judge’s verdicts, we obtain an F1 score of 0.921 (Precision of 0.953 and Recall of 0.891), indicating strong agreement between the LLM judge and human annotation on this binary presence-verification task. 

Computing this indicator at every session for every reference memory 
is nonetheless prohibitively expensive. We therefore approximate the 
inner summation by sampling $K{=}20$ evenly spaced checkpoints from 
$[t_\text{start}(r),\, T_\text{target}(r)]$—always including the 
first and last sessions—and rescale the sampled scores to approximate the full sum,
with the rescaling factor $\frac{|S(r)|}{K}$ treating each 
sampled checkpoint as representative of $\frac{|S(r)|}{K}$ 
consecutive sessions.
Full implementation details are provided in 
Appendix~\ref{sec:ap-rr-implementation}.

\subsection{Downstream Task Evaluation: Retrieval and QA}

While memory retention measures whether a memory system preserves reference information over time, it does not directly assess whether the retained information can be retrieved and used by an agent. We therefore complement retention evaluation with \emph{probing-question-based} downstream task evaluation. During life-skeleton construction, we associate each reference memory with a probing question whose answer depends on that memory. These questions cover both user-specific facts, such as preferences and traits, and task-state facts, such as decisions, milestones, and plans. 

This formulation provides a task-agnostic proxy for downstream utility, as these are precisely the types of information that a downstream agent must recover before it can correctly perform a task grounded in that information. For example, a personalized agent system cannot suggest an appropriate dish without knowing the user’s dietary preferences, while it cannot correctly continue a project without knowing its latest state or prior decisions. Correctly answering the associated probing question therefore tests the shared prerequisite of accessing the information needed for such downstream behavior, without tying the evaluation to a particular application or action space. 

Crucially, we conduct this downstream evaluation in the same evaluation loop as memory retention. For each reference memory $r$ and each sampled checkpoint $t \in S(r)$ used to estimate RR, we first apply the retention judge described above to the memory bank state $\mathcal{M}_t$. In the same pass, we use the probing question associated with $r$ to query $\mathcal{M}_t$, retrieve the top-$5$ entries, and prompt the LLM to answer the question using only those entries. Thus, retention, retrieval, and QA are all evaluated on the identical $(r, \mathcal{M}_t)$ instances and sampled checkpoints. \par Retrieval Accuracy records whether $r$ is included among the top-$5$ entries retrieved by its probing question. QA Accuracy records whether the generated answer is judged correct with respect to $r$. Together with the retention verdict, these measures distinguish whether a reference memory is preserved in the bank, accessible through retrieval, and usable for answering a downstream query.

\section{Can Memory Systems Be Personalized?}

An ideal personalized memory system should infer user-specific worth-storing information from interaction history and manage its memory bank accordingly. In this section, we conduct an empirical study to evaluate whether current LLM-based memory systems can be effectively personalized, and identify directions for future development.

\subsection{Experimental Setup}
\subsubsection{Memory Personalization via Session-level Storage Gating}

We propose a simple yet general framework for personalizing memory systems via \textit{session-level storage gating}. After each session, a gating module inspects the session dialogue and, optionally, prior context to predict whether the session is part of a long-horizon task or a transient interaction. If the session is classified as transient, memory operations for that session are skipped entirely. This lightweight wrapper requires no modification to the underlying memory system, and allows the memory budget to be concentrated on sessions that genuinely benefit from long-term context accumulation.

We evaluate gating methods along a spectrum of increasing contextual richness, from purely session-local signals to explicit structural modeling of cross-session dependencies. To bracket the range of achievable performance, we also define two reference points:

\textbf{Universal.} \@  The memory system operates without any gating, applying its default storage policy uniformly to all sessions. This represents the current state of deployed memory systems.

\textbf{Oracle.} \@ The ground-truth agent-use profile is provided directly, giving perfect knowledge of which sessions are long-horizon and which are transient. This serves as the upper bound for any gating method.

The three gating methods we evaluate are as follows.

\textbf{Greedy.} \@ The simplest instantiation of storage gating. At each session, an LLM predicts whether the session is long-horizon or transient based solely on the current session's dialogue, with no access to prior context. This captures the intuition that long-horizon and transient interactions often exhibit distinguishable surface-level patterns (e.g., references to ongoing goals vs.\ self-contained queries), without requiring any cross-session reasoning.

\textbf{Context-aware.} \@ To address the absence of historical signal in the Greedy method, each session is summarized in one to two sentences after processing, and a sliding window of the most recent $K$ summaries is passed as context when predicting subsequent sessions. This allows the gating module to exploit sequential patterns in the user's interaction history, particularly for sessions that are ambiguous in isolation.

\textbf{Structure-aware Method.} \@ Rather than treating history as a flat sequence of summaries, this method explicitly models the \textit{relational structure} among sessions---identifying which sessions form coherent long-horizon projects and which are isolated one-off interactions. To this end, we maintain a \textit{structural note}, a structured representation of the user's emerging usage patterns updated every $K$ sessions:
\begin{equation}
\small
\{
  \texttt{projects}: [\{
    \texttt{project\_id}, 
    \texttt{topic}, 
    \texttt{session\_ids}, 
    \texttt{status}
  \}],\;
  \texttt{isolated\_sessions}: [\texttt{session\_ids}]  \nonumber
\}
\end{equation}
Crucially, the note is carried forward across windows rather than reset, allowing the system to retroactively reassign sessions (e.g., recognizing that a previously isolated session belongs to a project identified later). Sessions in \texttt{isolated\_sessions} are classified as transient; sessions in any project are classified as worth storing. Among the three methods, Structure-aware is the only one that approximates domain-level usage pattern inference---the other two operate purely at the session level.

Implementation details for all methods are provided in Appendix~\ref{sec:ap-personalization-methods}.

\subsubsection{Memory Systems}

We adopt three recent memory systems as evaluation targets: Mem0~\cite{chhikara2025mem0}, Memory-R1~\cite{yan2025memoryr1}, and RMM~\cite{tan2025prospect}. These systems employ LLM-based memory operations—including selective extraction, storage, update, and deletion—to maintain a persistent memory bank. Memory operations are applied at two granularities: \textit{turn-level} and \textit{session-level}.\footnote{RMM supports session-level operation only.} For session-level settings, we use \texttt{gpt-5-mini} as the backbone LLM. For turn-level settings, where the higher frequency of LLM calls makes large models cost-prohibitive, we use the open-source \texttt{Qwen3-14B}.\footnote{As Memory-R1 does not release its trained model weights, we use the base LLM for this system.}

Memory entries are stored as text with associated embeddings in a vector database, and the memory budget is defined as the maximum number of entries allowed. To manage budget constraints, we adopt a hybrid deletion strategy combining each system's built-in mechanism with a rule-based time-decay approach following \cite{packer2023memgpt, wang2025m+, xu2025amem, hu2025memorysurvey}, where the oldest entries are evicted first to emulate human memory fading. Unless otherwise stated, all 
experiments use a default budget of 200 entries.

\subsubsection{Evaluation Metrics}

For session-wise gating quality, we report \textbf{F1}, \textbf{False Negative 
Rate (FNR)}, and \textbf{False Positive Rate (FPR)}. A false negative (FN) occurs 
when a worth-storing session is misclassified as transient. A false positive (FP) 
occurs when a transient session is stored unnecessarily. Both error types are 
undesirable, making F1 the primary overall measure.

For memory system performance, we use the Memory Retention Rate ($RR$), Retrieval Accuracy, and QA Accuracy, defined in \Cref{sec:evaluation-protocol}.

\subsection{Experimental Results}

\begin{figure*}[t]
\centering
\begin{minipage}{0.5\linewidth}
\centering
\captionof{table}{Session gating classification performance on \proposeddata$^s$ and \proposeddata$^d$. }
\label{tab:prediction-results}
\resizebox{\linewidth}{!}{%
\begin{tabular}{llcccccc}
\toprule
& & \multicolumn{3}{c}{\proposeddata$^s$} 
  & \multicolumn{3}{c}{\proposeddata$^d$} \\
\cmidrule(lr){3-5} \cmidrule(lr){6-8}
\textbf{Model} & \textbf{Method} 
  & F1 ($\uparrow$) & FNR ($\downarrow$) & FPR ($\downarrow$)
  & F1 ($\uparrow$) & FNR ($\downarrow$) & FPR ($\downarrow$) \\
\midrule
\multirow{3}{*}{\shortstack{Qwen3 \\ 14B}}
  & Greedy    & 0.660 & 0.457 & 0.117 & 0.657 & 0.444 & 0.178 \\
  & Context   & 0.434 & 0.715 & \textbf{0.008} 
              & 0.477 & 0.665 & \textbf{0.071} \\
  & Structure & \textbf{0.844} & \textbf{0.115} & 0.280 
              & \textbf{0.805} & \textbf{0.110} & 0.461 \\
\midrule
\multirow{3}{*}{\shortstack{gpt-5 \\ mini}}
  & Greedy    & 0.733 & 0.301 & 0.259 & 0.715 & 0.287 & 0.378 \\
  & Context   & 0.751 & 0.287 & \textbf{0.211} 
              & 0.733 & 0.261 & \textbf{0.375} \\
  & Structure & \textbf{0.795} & \textbf{0.018} & 0.652 
              & \textbf{0.784} & \textbf{0.010} & 0.782 \\
\bottomrule
\end{tabular}}
\end{minipage}
\hfill
\begin{minipage}{0.48\linewidth}
\centering
\includegraphics[width=\linewidth]{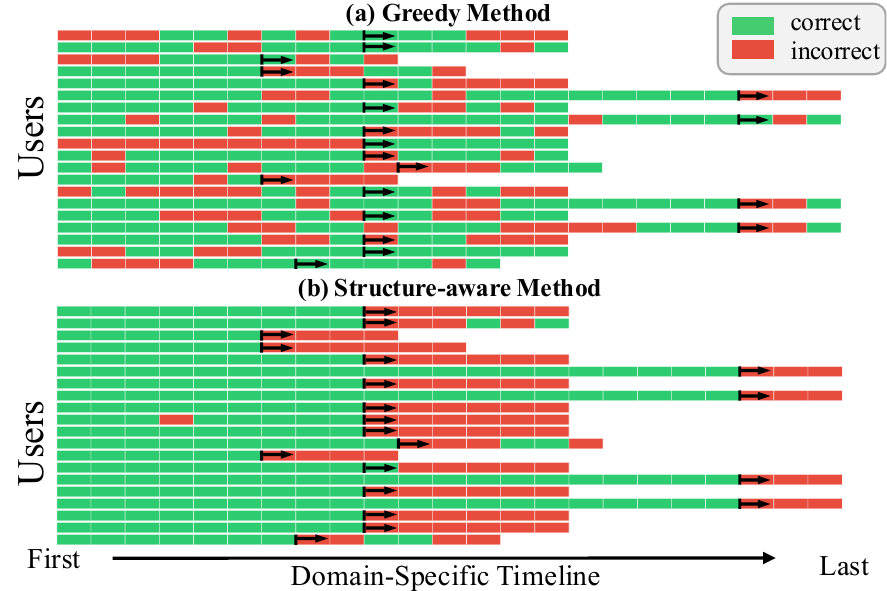}
\captionof{figure}{Session gating accuracy under a use profile shift. ``$\rightarrow$'' marks the shift point.}
\label{fig:transition}
\end{minipage}
\vspace{-3ex}
\end{figure*}

\textbf{Finding 1: Structure-aware gating shows promise in recovering domain-level usage 
patterns, but struggles to detect profile shifts.}

\begin{wrapfigure}{t}{0.6\linewidth}
    \centering
    \vspace{-4ex}
    \includegraphics[width=\linewidth]{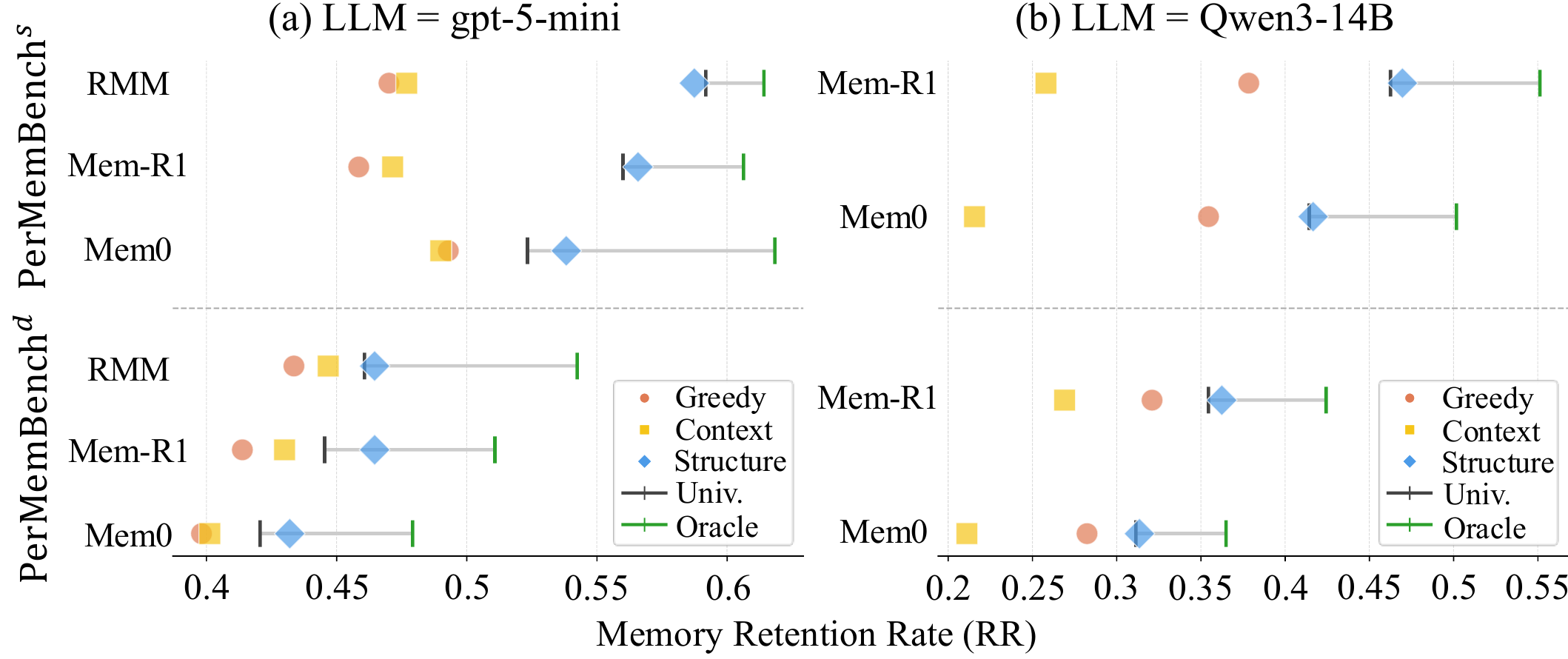}
    \vspace{-3ex}
    \caption{Memory storage performance comparison.}
    \vspace{-2ex}
    \label{fig:memory-performace}
\end{wrapfigure}
As shown in \Cref{tab:prediction-results}, the Structure-aware method achieves up to $0.844$ F1 on \proposeddata$^s$, demonstrating that explicitly modeling cross-session relational structure enables LLMs to approximate domain-level usage patterns from interaction history alone. 
In contrast, the lower performance of Greedy and Context-aware methods highlights that session-local signals — whether from the current session alone or a flat window of past summaries — are insufficient for this purpose, as these methods operate purely at the session level without inferring any domain-level structure.

\begin{wrapfigure}{t}{0.6\linewidth}
    \centering
    \vspace{-3ex}
    \includegraphics[width=0.95\linewidth]{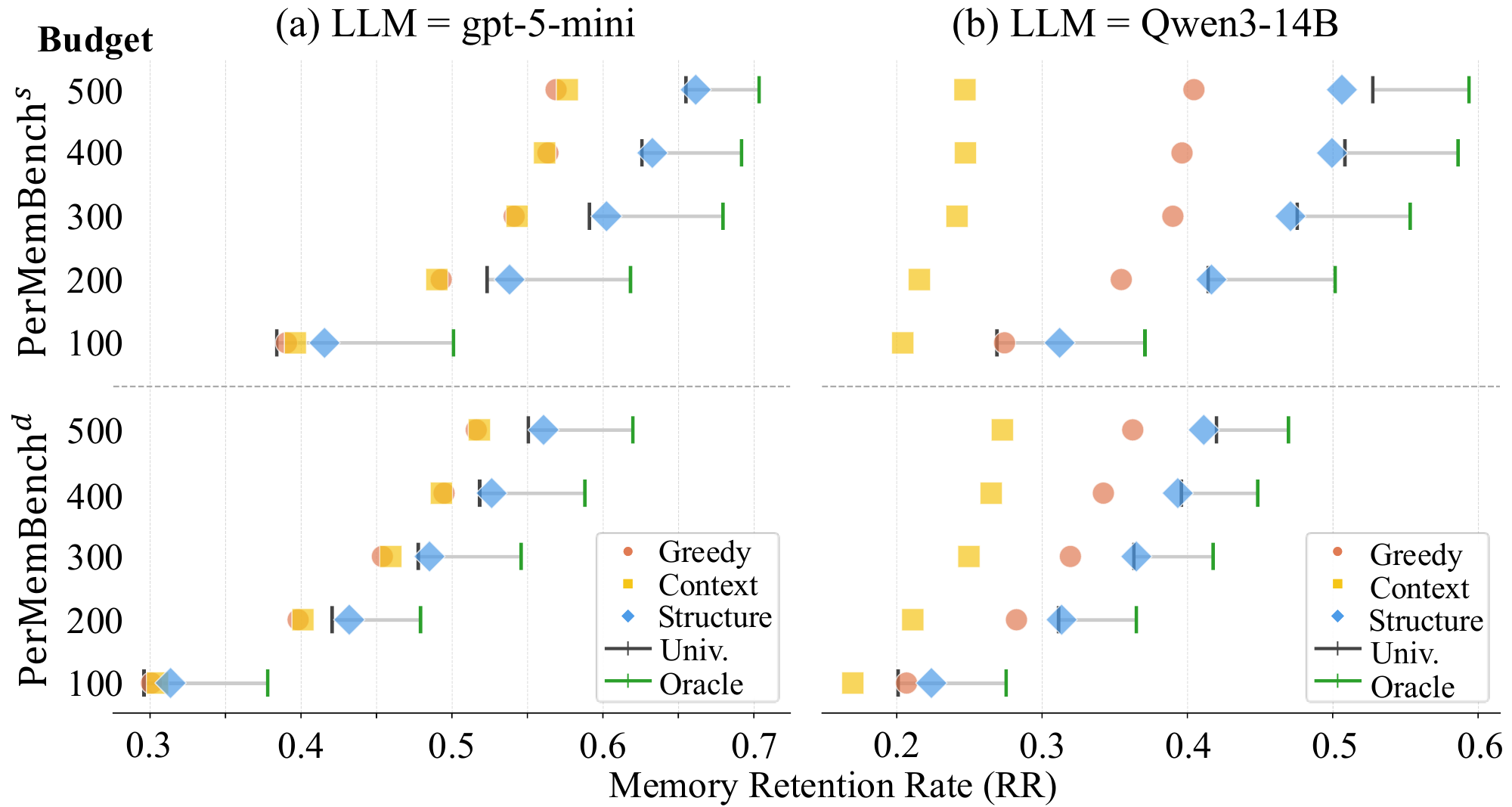}
    \vspace{-1ex}
    \caption{Sensitivity Analyses of memory budget on Mem0. The memory retention rate is reported.}
    \vspace{-4ex}
    \label{fig:sensitivity-budget}
\end{wrapfigure}
However, as shown in \Cref{fig:transition}, when we analyze prediction accuracy on domains undergoing a "long-horizon $\to$ transient shift", Structure-aware predictions are largely correct prior to the shift but collapse immediately afterward. We attribute this to over-reliance on established project structure: once a project cluster is formed in the structural note, the model continues assigning subsequent sessions to it even after the usage pattern has fundamentally changed. Interestingly, Greedy — which evaluates each session in isolation — shows no such collapse, exhibiting consistent performance before and after the shift. This suggests that combining the complementary strengths of structure-aware and session-level reasoning is a promising direction for robust session-level gating  under dynamic conditions.

\textbf{Finding 2: Memory personalization improves both retention and downstream utility over universal policy, with larger gains under tighter memory budgets.} 

As shown in \Cref{fig:memory-performace}, comparing Oracle and Universal across all memory systems reveals substantial retention improvements when the agent use pattern is known exactly. By avoiding wasteful storage on transient sessions and allocating the full budget to genuinely worth-storing contexts, personalization dramatically improves the utilization of limited memory capacity. Furthermore, \Cref{fig:sensitivity-budget} shows that the benefit of personalization is most pronounced at smaller budgets: the Oracle–Universal gap is largest at budgets of 100 and 200, and narrows as the budget increases. Nevertheless, even at a budget of 500, personalization continues to yield meaningful gains, confirming that the benefit is not limited to severely constrained settings.

\begin{wrapfigure}{t}{0.6\linewidth}
    \centering
    \vspace{-2ex}
    \includegraphics[width=\linewidth]{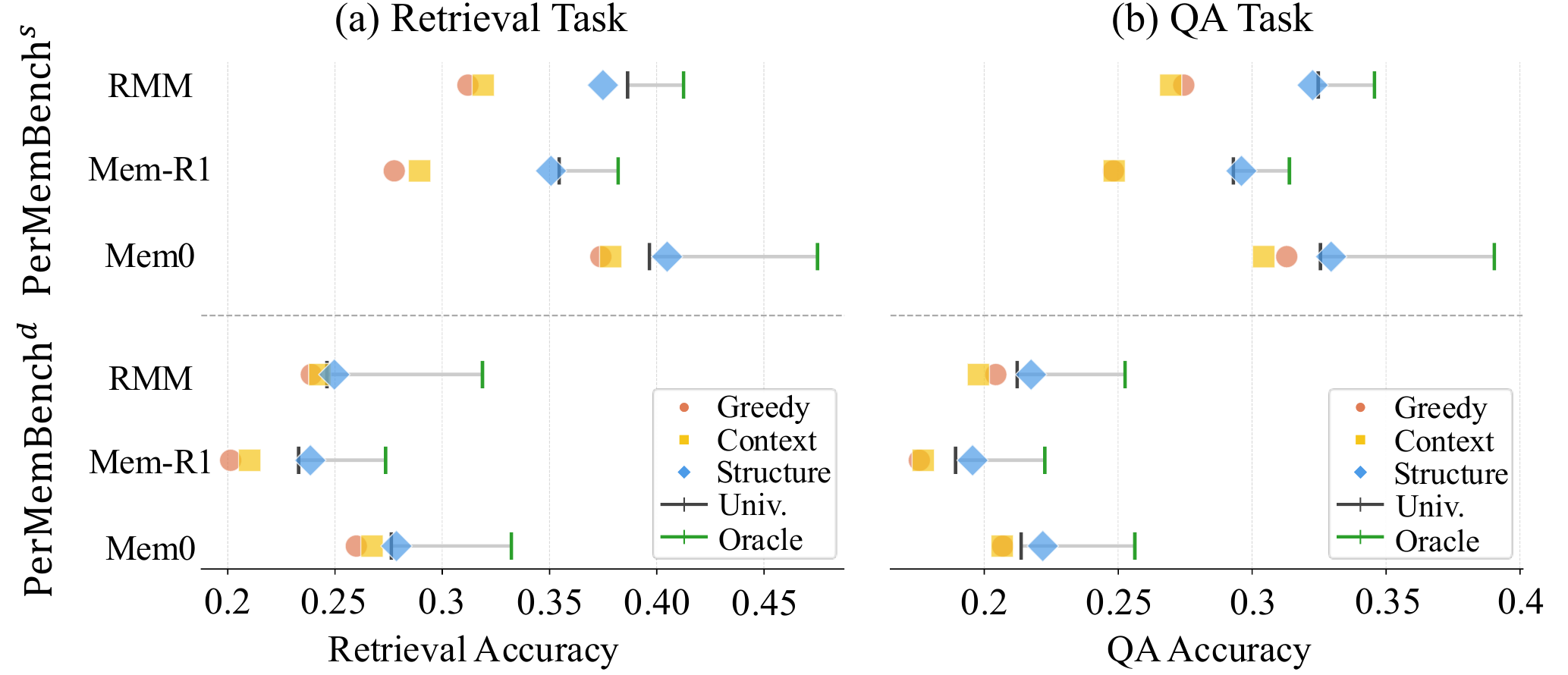}
    \caption{Personalized memory system performance (Retrieval and QA). For the backbone LLM, gpt-5-mini is used.}
    \vspace{-2ex}
    \label{fig:down-performace}
\end{wrapfigure}
This benefit extends beyond retention to downstream utility. As shown in \Cref{fig:down-performace}, Oracle consistently outperforms Universal on both Retrieval and QA across all memory systems and benchmark settings. Thus, when the agent-use profile is known, selectively allocating memory capacity to long-horizon sessions not only preserves more reference memories, but also makes the relevant information more likely to be retrieved and correctly used to answer task-relevant questions. Together with the retention results in \Cref{fig:memory-performace}, these findings establish that the potential benefit of personalization is functional rather than merely representational.

\textbf{Finding 3: The potential of personalization is large, but current gating performance is not yet accurate enough to realize it.}

However, when session-level gating is imperfect, the picture changes dramatically: Greedy and Context-aware consistently underperform Universal, and Structure-aware yields only marginal improvements despite its higher gating accuracy. The gap between Oracle and the best gating method thus represents unrealized potential — not an argument against personalization, but a 
precise measure of how much headroom remains. Closing this gap through more accurate session-level gating is therefore the most impactful direction for future work.

\section{Conclusion}
\vspace{-2ex}

We formalize the need for personalized memory systems and present \proposeddata, the first benchmark that jointly models multi-domain interaction histories and cross-user heterogeneity in agent-use patterns, along with a fully automated construction pipeline whose reliability is validated through rigorous meta-evaluation. We further propose a novel memory personalization paradigm, \textit{session-level storage gating}, along with simple baselines. Our results confirm that personalization yields substantial retention gains when the user's agent use pattern is exactly inferred, with benefits most pronounced under tighter memory budgets — underscoring the critical importance of memory personalization in resource-constrained deployment. Nevertheless, accurate session-level gating remains an open and pressing challenge for future work.

\section*{Acknowledgements}

This work was supported by National IT Industry Promotion Agency(NIPA) grant funded by the Korea government(MSIT) (RS-2026-25621604), IITP grant funded by the Korea government(MSIT) (RS-2026-25520654), and National Research Foundation of Korea(NRF) funded by Ministry of Science and ICT (RS-2022-NR068758).

{\small
\bibliographystyle{unsrt}
\bibliography{refer}
}


\newpage


\appendix

\section{Limitations and Future Works}

\textbf{Benchmark scale.} \@ \proposeddata~currently encompasses 20 users, which may limit the diversity of agent use patterns represented. We note, however, that our fully automated construction pipeline can be readily scaled to larger and more diverse user cohorts without manual intervention.

\textbf{Simplicity of agent use profile modeling.} \@ We model agent use profiles as a joint configuration of domain participation and memory necessity — a deliberately simple formulation. While finer-grained behavioral patterns undoubtedly exist, our primary goal is to shed the first light on the necessity of memory personalization and lay a foundational stepping stone for this research direction. Extending \proposeddata~to richer profile representations remains an important avenue for future work.

\textbf{Personalization limited to storage operations.} \@ Our proposed \textit{session-level storage gating} paradigm exclusively applies to storage, with no mechanism to retroactively correct mistakenly stored entries. A personalized deletion policy that evicts user-specifically unnecessary memories could make memory management substantially more effective, and we leave this as future work.

\textbf{Session-level gating accuracy.} \@ As shown in Finding 3, current gating methods remain insufficient to fully realize the gains of personalization. We believe agentic post-training — optimizing the gating module through agent-environment interaction — is a promising direction for closing this gap.

\textbf{Human-in-the-Loop Memory Personalization.} \@ An important direction for future work is to incorporate explicit user control into personalized memory management. Rather than allowing agents to autonomously determine which information should be retained, practical systems could adopt a user-driven memory personalization paradigm. Specifically, users could explicitly opt in to personalization during onboarding and review candidate memory entries extracted from previous interactions before approving their storage. Furthermore, a dedicated memory management interface could allow users to inspect, edit, correct, or delete stored memories at any time. Such mechanisms would improve transparency and user agency while mitigating the risks of unwanted profiling and incorrect memory retention.

\section{Societal Impacts and Risks}
\label{sec:societal_impacts}

While long-horizon personalized memory frameworks significantly enhance the continuity and helpfulness of AI agents, deploying persistent memory systems introduces non-trivial ethical, privacy, and safety risks that warrant careful consideration.

\paragraph{Surveillance Dangers and Behavioral Profiling.}
Silently tracking and aggregating user interactions across multiple life domains over long periods creates a detailed behavioral digital footprint. If mishandled, such long-term profiles could be exploited for intrusive surveillance, unwanted psychological profiling, or targeted behavioral manipulation. To mitigate these risks, memory personalization should be designed as a user-governed process rather than a fully automated and opaque one. In particular, systems should obtain explicit user consent before storing personal information, especially when the information is sensitive or intended to be retained for long periods.

\paragraph{User Control, Auditability, and Memory Correction.}
A static or opaque memory store risks retaining outdated, incorrect, or sensitive personal inferences. Such errors can lead to inappropriate personalization and may repeatedly reinforce incorrect assumptions about the user. To preserve user agency over personal data, memory systems should provide transparent and granular controls that allow users to inspect, edit, correct, and delete stored memories. They should also maintain auditable records of when, why, and from which interaction a memory was created or updated.

\paragraph{Cross-User Data Isolation and Leakage Safeguards.}
In multi-user or multi-tenant deployment environments, cross-user memory contamination represents a serious privacy risk. Although our benchmark strictly isolates the memory database of each synthetic persona, real-world systems may still be vulnerable to leakage through shared indexes, caches, retrieval pipelines, or backend infrastructure. Developers should therefore enforce strict tenant-level data boundaries, robust authentication and authorization, encryption, and systematic access verification. Production systems should also be regularly audited and tested to ensure that one user's stored history cannot be retrieved, inferred, or exposed in another user's interactions.

\section{Benchmark Construction Details}

\subsection{Domain Pool Construction}
\label{sec:ap-domain-pool}

To ensure representative coverage of real-world usage, we employ a data-driven approach to construct a domain pool. First, we sample 1,000 personas and prompt \texttt{Claude-Haiku-4.5} to generate potential usage scenarios without predefined constraints using following prompt:

\begin{promptbox}[Domain Extraction Prompt]
\small
You are analyzing user personas to understand how people use AI agents in their daily lives.

\textbf{User Persona}

\{\texttt{persona}\}

\textbf{Task}

Based on this user's background, lifestyle, occupation, hobbies, and goals,
infer what kinds of tasks or purposes this user would use an AI agent for.

\textbf{Instructions}

- Generate between 5 and 10 distinct domains of AI agent use for this user.

- Each domain must be grounded in this user's specific context. Do not generate generic domains that would apply to any user.

- domain\_name must be short (2-5 words) and general enough to apply to other users with similar backgrounds. All persona-specific context belongs in the reason field only.

  - Too specific (bad): "ML/NLP research ideation, experiment design, and literature synthesis"
  
  - Good: "Academic Research \& Literature Review"
  
  - Too vague (bad): "Work-related tasks"

- Cover a diverse range of usage purposes across different aspects of the persona (work, hobbies, health, personal goals, etc.). Do not cluster domains around a single aspect of the persona.

\textbf{Output Format}

Return a JSON object in the following format:

[
    \{
    
      "domain\_name": "...",
      "reason": "One sentence grounding this domain in this user's specific context."
      
    \}
  ]
  
\end{promptbox}

These candidates are then semantically clustered and assigned representative labels via human review. To align the pool with actual LLM trends, we cross-reference these clusters with industry reports \cite{chatterji2025people, openai2026chatgpt_usage_adoption_work}, pruning niche cases and supplementing broad-interest domains. This process results in a final taxonomy of 20 domains (see \Cref{tab:added_domains}).

\begin{table}[h]
\centering
\caption{The list of defined domain pools.}
\begin{tabular}{cl}
\toprule
\textbf{\#} & \textbf{Domain} \\
\midrule
1& Academic Study \& Learning \\
2&Business \& Entrepreneurship \\
3& Career Development \& Job Search \\
4& Data Analysis \& Visualization \\
5& Event Planning \\
6& Health \& Wellness \\
7& Home \& Real Estate  \\
8& Language Learning \\
9& Legal \& Administrative Affairs \\
10& Math \& Quantitative Problem Solving \\
11& Mental Health \& Emotional Support \\
12&News \& Current Events \\
13& Personal Finance \& Investment \\
14& Recipe Advice \& Meal Planning \\
15&Relationship \& Social Advice \\
16&Shopping \& Product Research \\
17& Software Development \& Coding \\
18& Sport \& Physical Activity \\
19 & Travel Planning \\
20 & Writing Assistant \\
\bottomrule
\end{tabular}
\label{tab:added_domains}
\end{table}

\subsection{User-Specific Profile Assignment}
\label{sec:ap-profile-assign}

For each persona $p \in \mathcal{P}$ and domain $d \in \mathcal{D}$, we employ \texttt{Claude-Haiku-4.5} to infer profiles based on the user’s lifestyle and objectives using the following prompt:

\begin{promptbox}[User-Specific Profile Assignment Prompt]
\scriptsize
You are building a user profile for a personalized AI agent system.

\texttt{\#\# User Persona}

\{persona\}

\texttt{\#\# Domains}

\{domain\_list\}

\texttt{\#\# Task}

For every domain listed above, determine whether this user would use an AI agent for it, and if so, whether memory is required and how frequently they would use it.

\texttt{\#\# Instructions}

- Evaluate all {n\_domains} domains without exception.

- For each domain, first determine whether this user would plausibly use an AI agent for it (use: true/false).

- If use is true, determine whether memory is required (memory\_required: true/false) and how frequently this user would use this domain (frequency: "high"/"medium"/"low").

- If use is false, set memory\_required to null and frequency to null.

- Base your judgment entirely on this user's specific context, not on general assumptions about the domain.

\texttt{\#\# Definition of Memory Required}

Memory is required (true) when this user's usage of the domain is:

- Ongoing and accumulative (e.g., tracking progress over time)

- Connected across multiple sessions (e.g., building on past conversations)

- Tied to long-term goals or evolving personal circumstances

Memory is NOT required (false) when this user's usage of the domain is:

- One-time or ad-hoc (e.g., a single lookup with no follow-up)

- Self-contained within a single session

- Not dependent on past interactions

\texttt{\#\# Definition of Frequency}

- "high": This user would use an AI agent for this domain very regularly

- "medium": This user would use an AI agent for this domain occasionally

- "low": This user would use an AI agent for this domain rarely

\texttt{\#\# CRITICAL}

Both memory\_required and frequency must reflect THIS USER's specific context, not the general nature of the domain.

The same domain can have different memory\_required and frequency values for different users.

For example:

- "Recipe Advice \& Meal Planning" → memory\_required=false, frequency="low" for a user who occasionally looks up recipes, but memory\_required=true, frequency="high" for a user who is developing their own menu or working toward opening a restaurant.

- "Fitness \& Exercise Planning" → memory\_required=false, frequency="low" for a user who casually checks workout tips, but memory\_required=true, frequency="high" for a user who follows a structured training routine and tracks progress over time.

- "News \& Current Events" → memory\_required=false, frequency="low" for a user who reads news casually, but memory\_required=true, frequency="high" for a user who actively monitors specific topics for research or professional purposes.

\texttt{\#\# Output Format}

Return a JSON object in the following format:
\begin{verbatim}
[
    {
      "domain_name": "...",
      "use": true/false,
      "memory_required": true/false/null,
      "frequency": "high"/"medium"/"low"/null,
      "reason": "One sentence explaining why, grounded in this user's specific context."  
    }
  ]
\end{verbatim}
\end{promptbox}

\subsection{Life Skeleton and Timeline Construction Details}
\label{sec:ap-skeleton-timeline}

This subsection provides a detailed description of the Life Skeleton and Timeline 
Construction pipeline (\S\ref{sec:life-timeline-construction}), covering both 
\proposeddata$^s$ (static) and \proposeddata$^d$ (dynamic).

\subsubsection{Life Skeleton Construction for \proposeddata$^s$}
\label{sec:ap-phase1-skeleton}

\paragraph{Long-Horizon Domains.}
For each memory-required domain ($m_{p,d}=1$), an LLM generates a structured life skeleton comprising a sequence of projects and events. The number of projects and events per project is determined by the frequency metadata $f_{p,d}$ as follows:

\begin{center}
\small
\begin{tabular}{lccc}
\toprule
\textbf{Frequency} & \textbf{\# Projects} & \textbf{Events/Project} \\
\midrule
High   & 5 & 3--5 \\
Medium & 3 & 2--4 \\
Low    & 2 & 2--3 \\
\bottomrule
\end{tabular}
\end{center}

Each event is annotated with reference memory items of two types: \texttt{user\_profile} (stable facts persisting across all projects) and \texttt{ongoing\_state} (project-scoped decisions and progress). To prevent redundant reference memories across domains, skeletons are generated sequentially — each domain receives a list of facts already recorded by previously generated domains, and is instructed not to duplicate them.

\begin{promptbox}[Life Skeleton Generation (Memory-Required Domain) Prompt (condensed version)]
\small
\textbf{[System]}\\
You are designing a Life Skeleton for a personalized AI agent memory benchmark.
A Life Skeleton captures how a specific user engages with an AI agent in a particular domain over 1--2 years\ldots [see full prompt in code]

\textbf{[User]}\\
\texttt{\#\# User Persona}\\
\{persona\}

\texttt{\#\# Domain}\\
- Name: \{domain\_name\}  Frequency: \{frequency\} \quad Why used: \{reason\}

\texttt{\#\# Task}\\
Generate a Life Skeleton over a 1--2 year period.
Scale: \{n\_projects\} projects, \{n\_events\_min\}--\{n\_events\_max\} events per project.

\texttt{\#\# GT Memory Definitions}\\
\textbf{user\_profile} — persists forever after learned.
Skills, tools mastered, revealed preferences. Only record if NOT already in persona.

\textbf{ongoing\_state} — lives only while the project is active.
Decisions made, tools chosen, progress reached. 
CRITICAL: Must be things decided DURING conversation — not things user already knows.

\texttt{\#\# Already Covered (DO NOT duplicate)}\\
\{covered\_facts\_from\_prior\_domains\}

Output: JSON with \texttt{projects[].events[].gt\_memory[\{type, fact, probing\_question, answer\}]}
\end{promptbox}

\paragraph{Transient Domains.}
For transient domains ($m_{p,d}=0$), independent one-off events are generated without project structure or reference memory. The number of events is derived from the timeline duration estimated from the memory-required skeletons and the domain's interaction frequency:
\begin{equation}
n_{\text{events}} = \left\lfloor \frac{T_{\text{total}} \times 4}
{w_{f_{p,d}}} \right\rfloor
\end{equation}
where $T_{\text{total}}$ is the estimated timeline duration in months 
and $w_{f}$ is the inter-session interval in weeks 
($w_{\text{high}}=4$, $w_{\text{medium}}=8$, $w_{\text{low}}=12$).

\begin{promptbox}[One-Off Event Generation (Transient Domain) Prompt (condensed version)]
\small
\textbf{[System]}\\
You are designing one-off AI agent interaction events\ldots
These events are self-contained, single-session interactions with no longitudinal engagement, no project structure, and no memory required across sessions.

\textbf{[User]}\\
\texttt{\#\# User Persona}\\
\{persona\}

\texttt{\#\# Domain}\\
- Name: \{domain\_name\} \quad Frequency: \{frequency\}

\texttt{\#\# Task}\\
Generate exactly \{n\_events\} one-off events spread across $\sim$\{total\_months\} months. Each event should reflect a genuinely different moment and need.

Output: JSON with \texttt{events[\{event\_id, event\_title, 
event\_description\}]}
\end{promptbox}

\subsubsection{Timeline Integration for \proposeddata$^s$}
\label{sec:ap-phase1-timeline}

Once per-domain skeletons are constructed, an LLM arranges all memory-required events into a unified chronological timeline. The LLM is responsible solely for placing memory-required events; 
transient events are placed programmatically after the LLM call using the frequency-based spacing formula above. Events beyond the LLM-determined \texttt{total\_months} are silently truncated.

\begin{promptbox}[Timeline Arrangement Prompt (condensed version)]
\small
\textbf{[System]}\\
You are designing an integrated session timeline\ldots
Your job is to arrange all events into a single, realistic session timeline — a chronologically ordered list of agent sessions that this person would actually have, given their real life circumstances. 

\texttt{\#\# Key Principles}

1. Ground in the persona's life.

2. Respect project sequentiality within each domain.
   
3. Interleave domains realistically.

4. Assign concrete month numbers.
   
5. Flag cross-domain links.

\textbf{[User]}\\
\texttt{\#\# User Persona}\\
\{persona\}

\texttt{\#\# Domain Life Skeletons}\\
\{skeleton\_summary\}

\texttt{\#\# Events to Place (all must appear exactly once)}\\
\{domain | project\_id | event\_id | event\_title\}

\texttt{\#\# Requirements}\\
- Every event must appear exactly once.\\
- Assign realistic month (1--\{max\_month\}) to each session.\\
- Identify anchor life events triggering multiple domains simultaneously.\\
- Respect project ordering within each domain.\\
- session\_id must be sequential in chronological order.

Output: JSON with \texttt{total\_months, anchor\_life\_events, session\_sequence}
\end{promptbox}

\subsection{Profile Shift and Life Skeleton Construction for \proposeddata$^d$}
\label{sec:ap-phase2-skeleton}

\proposeddata$^d$ extends \proposeddata$^s$ by introducing a profile shift at the end of the timeline in \proposeddata$^s$. The shift is determined by three fixed rules applied via rule-based sampling, with no LLM involvement:

\begin{enumerate}[leftmargin=0.5cm]
  \item \textbf{Demotion}: one existing memory-required domain is demoted to transient.
  \item \textbf{New longitudinal domain}: one new memory-required domain is sampled from the unused pool.
  \item \textbf{New transient domain}: one new transient domain is optionally added from the unused pool.
\end{enumerate}

Domains are sampled with frequency-weighted probability ($w_{\text{high}}=3$, $w_{\text{medium}}=2$, $w_{\text{low}}=1$) and each persona receives a deterministic seed derived from the global seed and its UUID, ensuring reproducibility.

\paragraph{Transition Narrative.}
A coherent life transition event is generated to justify all three changes simultaneously.

\begin{promptbox}[Life Transition Narrative Generation Prompt (condensed version)]
\small
\textbf{[System]}\\
You are writing a life transition narrative\ldots
Domain changes have already been decided. Write a short, coherent 
life transition event that naturally explains all the changes.

\textbf{[User]}\\
\texttt{\#\# User Persona}: \{persona\}

\texttt{\#\# Current Usage (Phase 1, \{total\_months\} months)}\\
Memory-Required: \{mem\_domains\} \quad One-Off: \{oneoff\_domains\}

\texttt{\#\# Phase 2 Changes (already decided)}\\
1. DEMOTED to occasional use: \{mem\_to\_oneoff\_name\}\\
2. NEW longitudinal domain: \{added\_mem\_name\}\\
3. NEW one-off domain: \{added\_oneoff\_name\}

Output: JSON with \texttt{\{name, description\}}
\end{promptbox}

\paragraph{Life Skeleton Generation after the Shift.}
Using the transition narrative, Phase 2 skeletons are generated for each domain with two variants: \textit{added} (new domains starting from scratch) and \textit{retained} (existing domains continuing into Phase 2 with new projects). The demoted domain is treated as a transient domain and generates one-off events instead. Prompts follow the same structure as in \proposeddata~generation pipeline, with the transition event appended as additional context and a list of reference memories in \proposeddata$^s$ to avoid duplication.

\paragraph{Timeline Integration after the Shift.}
Events occurring after the shift are arranged into a timeline using the same LLM prompt structure as previous one, with months expressed relative to the point where the shift starts. 
The final output is a continuous \texttt{all\_sessions} list spanning both phases.

\subsection{Dialogue Generation via Dual-Simulator}
\label{sec:ap-dialogue-generation}

Each entry in the unified timeline corresponds to one dialogue session, 
generated by two separate LLM instances: a \textit{user simulator} and 
an \textit{agent simulator}. The two simulators are strictly isolated — 
the agent simulator receives no access to the life skeleton or reference 
memories, responding solely based on the user's utterances and its 
parametric knowledge, exactly as a real deployed agent would.

Sessions are generated differently based on \texttt{memory\_required}: 
memory-required sessions ($m_{p,d}=1$) provide the user simulator with 
the event description and reference memory items from the life skeleton, 
while transient sessions ($m_{p,d}=0$) provide only the event description 
with no reference memory, terminating once the one-off need is met.

A key design challenge is preventing the user simulator from explicitly 
declaring reference memory facts upfront, which would make the resulting 
dialogues artificially unnatural. To address this, \texttt{user\_profile} 
facts are framed as background traits that should surface implicitly 
through the user's reactions, while \texttt{ongoing\_state} facts are 
converted into open uncertainties before being passed to the simulator — 
so that decisions emerge organically during the conversation rather than 
being pre-announced. Cross-session continuity is maintained by providing 
the user simulator with a summary of facts already established in prior 
sessions of the same project.

To verify that reference memory facts are covered within each session, 
an LLM judge tracks which facts have been expressed by the user after 
each turn, and surfaces any unrevealed facts as gentle nudges in the 
continuation prompt. The session concludes once all facts have been 
naturally revealed and a minimum turn count is reached.

The prompt structure provided to the user simulator is as follows.

\begin{promptbox}[User Simulator Prompt (condensed version)]
\small
\textbf{[System]}\\
You are roleplaying as a real person interacting with an AI agent.
\{persona\}. You are always the USER seeking help, never the assistant.

\textbf{[Opening turn]}\\
You are about to start a new conversation about: \{domain\_name\}

\texttt{\#\# What is happening in your life right now}\\
\{event\_description\}

\texttt{\#\# What you already know from previous conversations}\\
\{prior\_context\}

\texttt{\#\# Who you are in this conversation}\\
- \textit{[user\_profile]} Traits and preferences — let these shape 
  how you react. DO NOT announce them upfront.\\
- \textit{[ongoing\_state]} Things you have NOT yet decided — 
  let these emerge through the conversation.

Open with 1--3 sentences only. The agent has NO memory of you.

\textbf{[Continuation turn]}\\
\texttt{\#\# Not yet surfaced:} \{unrevealed\_facts\}\\
\texttt{\#\# Already came up — do NOT repeat:} \{revealed\_facts\}

Reply \texttt{[END]} if the conversation has reached a natural conclusion.
\end{promptbox}

\section{Meta Evaluation Details}
\label{sec:ap-meta-evaluation}

This section describes the detailed procedures used to validate the 
reliability of our data generation pipeline across three stages.

\subsection{Stage 1: Profile Plausibility}

We randomly sample 100 personas from the Nemotron-Persona-USA dataset 
and run our agent use profile assignment algorithm on each. 
The resulting profiles are then evaluated for plausibility by the 
two-evaluator panel. The prompt provided to \texttt{Claude Opus 4.6} 
is as follows.

\begin{promptbox}[Profile Plausibility Judge Prompt]
\small
You are a strict evaluator for persona-conditioned metadata quality.

You must decide whether the metadata for this user is overall valid and plausible, based on the user's persona.

\texttt{\#\# User Persona}

\{persona\_text\}

\texttt{\#\# Generated Metadata}

\{domains\_json\}

\texttt{\#\# Decision Criteria}

Judge "YES" only if ALL are true:

1) Domain-level choices are generally persona-grounded.

2) There are no major logical contradictions.

3) use/memory\_required/frequency are mostly coherent with each other.

4) Reasons are mostly plausible and consistent with the persona.

Judge "NO" if there are clear invalid/unreasonable patterns such as:

- repeated persona-irrelevant domains marked as use=true without evidence,

- frequent inconsistency (e.g., use=false but memory/frequency not null),

- clearly implausible frequency/memory decisions for many domains,

- major contradiction between reasons and persona.

\texttt{\#\# Output format (STRICT JSON only, no markdown)}
\begin{verbatim}
{
  "verdict": "YES" or "NO",
  "confidence": 0.0 to 1.0,
  "summary": "2-4 concise sentences",
  "major_issues": [
    "issue 1", 
    "issue 2"
  ]  
}
\end{verbatim}
\end{promptbox}

\begin{figure*}[h] 
    \centering
    \includegraphics[width=1\linewidth]{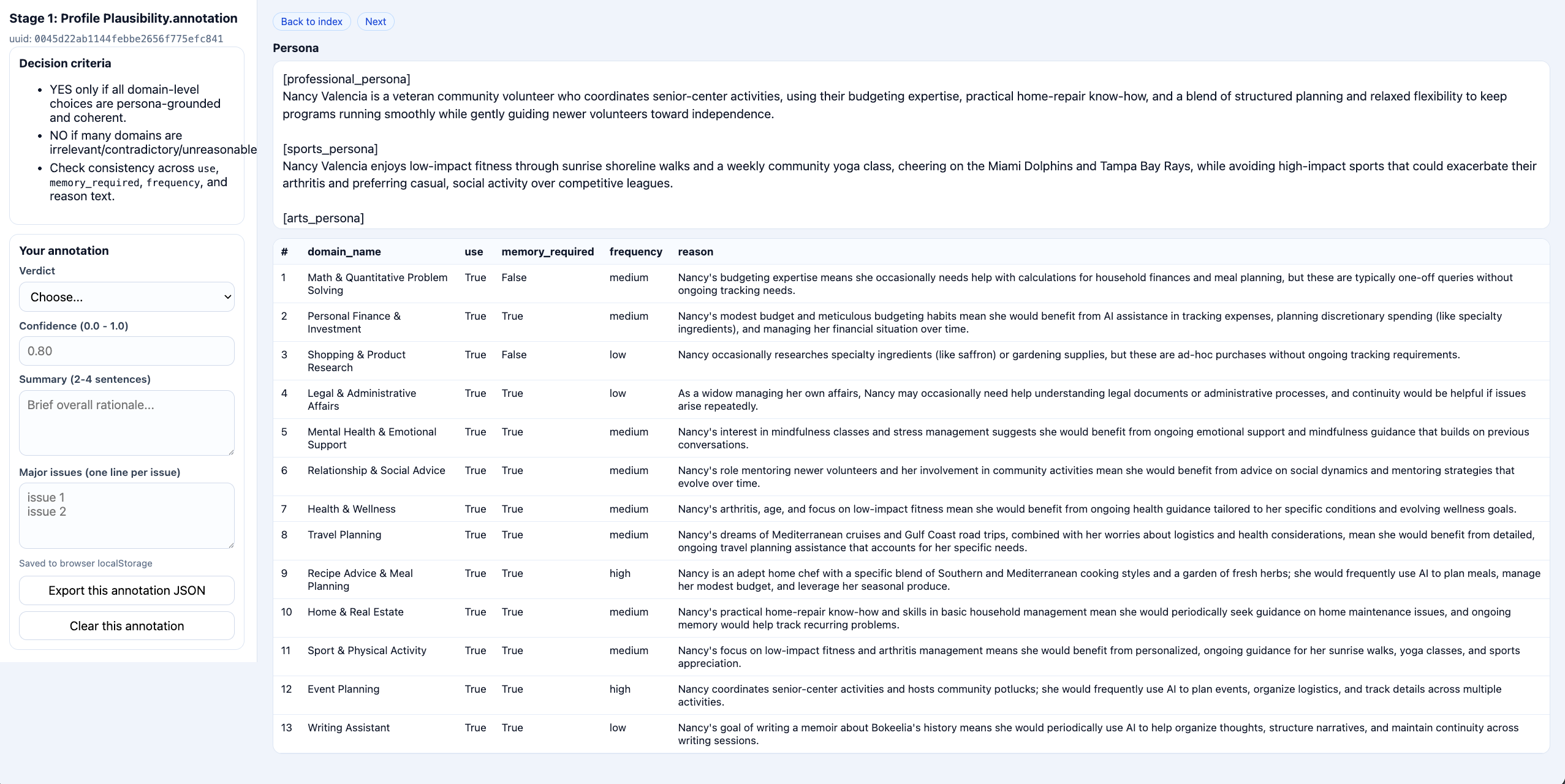}
    \caption{The user interface provided to the annotators for profile plausibility annotation.}
    \label{fig:profile-annotation-ui}
\end{figure*}

The annotation interface provided to the human expert is shown in \Cref{fig:profile-annotation-ui}.

\subsection{Stage 2: Life Skeleton and Timeline Realism}

Using the same 100 sampled personas, we run our life skeleton and 
timeline construction algorithm and evaluate the quality of the 
resulting outputs. The prompt provided to \texttt{Claude Opus 4.6} 
is as follows.

\begin{promptbox}[Life Skeleton and Timeline Realism Judge Prompt]
\small
You are a strict evaluator of synthetic life-skeleton quality.

Your task is to judge whether this user's generated life skeleton (especially project/event structure) is plausible and realistic.

\texttt{\#\# User Persona}

\{persona\_text\}

\texttt{\#\# Life Skeleton to Evaluate}

\{life\_skeleton\_json\}

\texttt{\#\# What to evaluate}

Focus on these criteria:

1) persona\_alignment:
   
   - Are projects/events grounded in this persona's background, constraints, habits, and goals?

2) project\_event\_structure:

   - Are project scopes and event granularity reasonable for real life?
   
   - Do events represent natural moments when user would open an AI agent?

3) timeline\_coherence:

   - Is there believable progression across projects/events?
   
   - Do durations and ordering feel coherent over 1-2 years?

4) realism\_plausibility:

   - Do scenarios feel realistic rather than template-like or contrived?
   
   - Are there major logical contradictions?

\texttt{\#\# Verdict rule}

- Return overall\_verdict = "YES" only if all 4 criteria are mostly satisfied and no major contradiction exists.

- Return overall\_verdict = "NO" if one or more criteria fail clearly.

\texttt{\#\# Output format (STRICT JSON only; no markdown)}
\begin{verbatim}
{
  "overall_verdict": "YES" or "NO",
  "confidence": 0.0 to 1.0,
  "criteria": {
    "persona_alignment": "YES" or "NO",
    "project_event_structure": "YES" or "NO",
    "timeline_coherence": "YES" or "NO",
    "realism_plausibility": "YES" or "NO"
  },
  "summary": "2-4 concise sentences",
  "major_issues": [
    "issue 1", 
    "issue 2"
  ]  
}
\end{verbatim}
\end{promptbox}

\begin{figure*}[h] 
    \centering
    \includegraphics[width=1\linewidth]{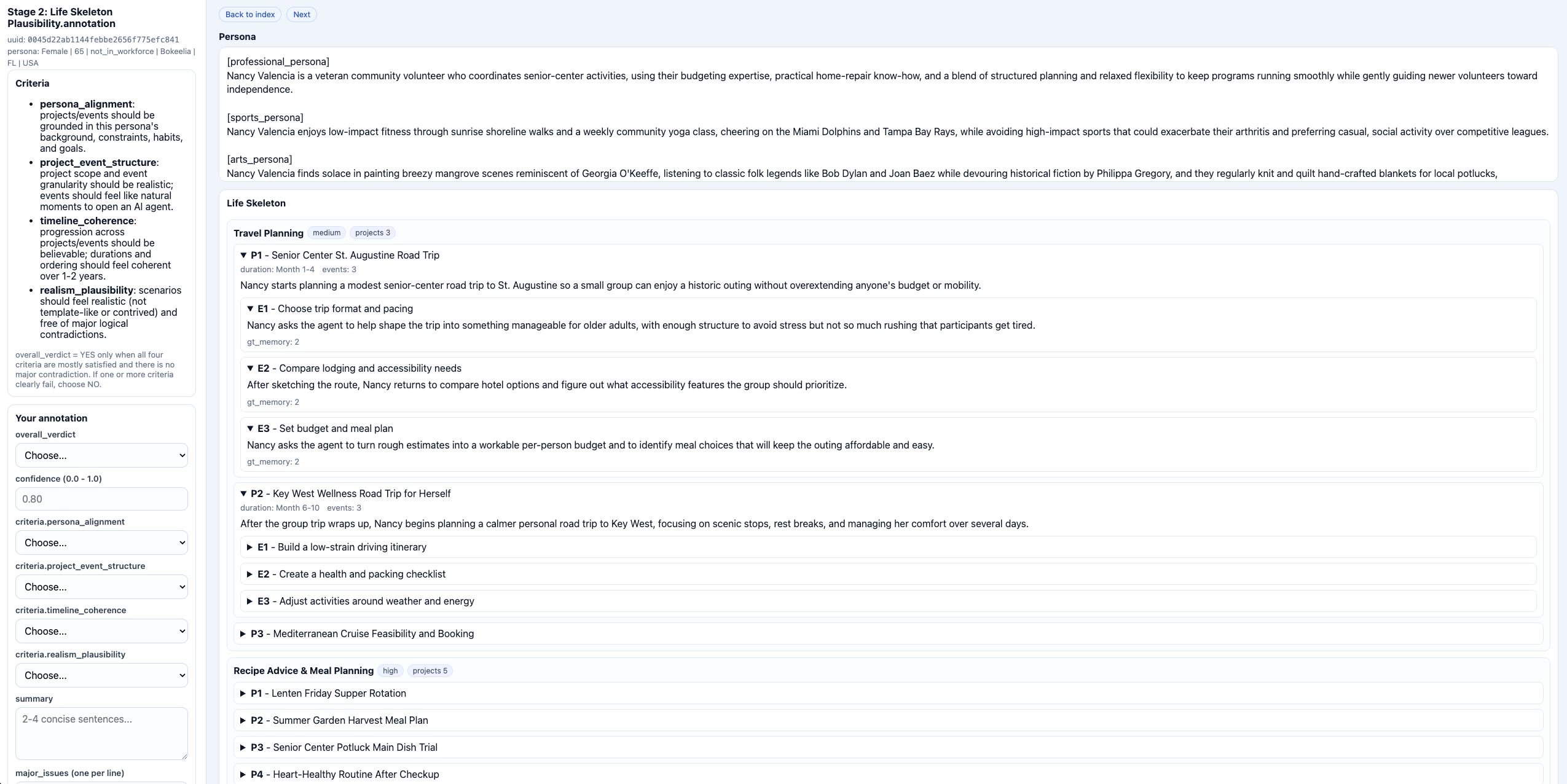}
    \caption{The user interface provided to the annotators for life skeleton and timeline realism annotation.}
    \label{fig:life-annotation-ui}
\end{figure*}

The annotation interface provided to the human expert is shown in \Cref{fig:life-annotation-ui}.

\subsection{Stage 3: Dialogue Quality}

Dialogues in our pipeline are stored at the session level, with each 
session corresponding to a single event in the life skeleton. 
Accordingly, we evaluate two aspects: whether the dialogue is 
consistent with the project and event context defined in the skeleton, 
and whether the reference memories defined for that event surface 
naturally during the user--agent interaction. We randomly sample 
100 sessions across users for evaluation. The prompt provided to 
\texttt{Claude Opus 4.6} is as follows.

\begin{promptbox}[Dialogue Quality Judge Prompt]
\small
You are a strict evaluator of synthetic user-agent dialogue quality.

You will evaluate one conversation generated from a life skeleton session.

\texttt{\#\# Session Context (ground truth)}

\{session\_context\}

\texttt{\#\# Conversation}

\{dialogue\_text\}

\texttt{\#\# Evaluation criteria}

1) skeleton\_alignment (1-5):

   - Does the dialogue align with the session's project/event context?
   
   - Is the user request trajectory consistent with event\_description?

2) gt\_memory\_revelation (1-5):

   - Do relevant gt\_memory facts surface naturally in conversation?
   
   - Avoid penalizing if only truly irrelevant facts are not surfaced.
   
   - Penalize forced, list-like, or unnatural fact dumping.

\texttt{\#\# Additional rule}

- overall\_verdict must be "YES" only if all three scores are at least 4 and there is no critical issue.

- Otherwise overall\_verdict must be "NO".

\texttt{\#\# Output format (STRICT JSON only; no markdown)}
\begin{verbatim}
{

  "overall_verdict": "YES" or "NO",
  "confidence": 0.0 to 1.0,
  "scores": {
    "skeleton_alignment": 1 to 5,
    "gt_memory_revelation": 1 to 5
  },
  "fact_checks": [
    {
      "fact": "original fact text", 
      "revealed": "YES" or "PARTIAL" or "NO",
      "evidence": "short quote or reason"
    }
  ],
  "summary": "2-4 concise sentences",
  "major_issues": [
    "issue 1",
    "issue 2"
  ]
}
\end{verbatim}
\end{promptbox}

\begin{figure*}[h] 
    \centering
    \includegraphics[width=1\linewidth]{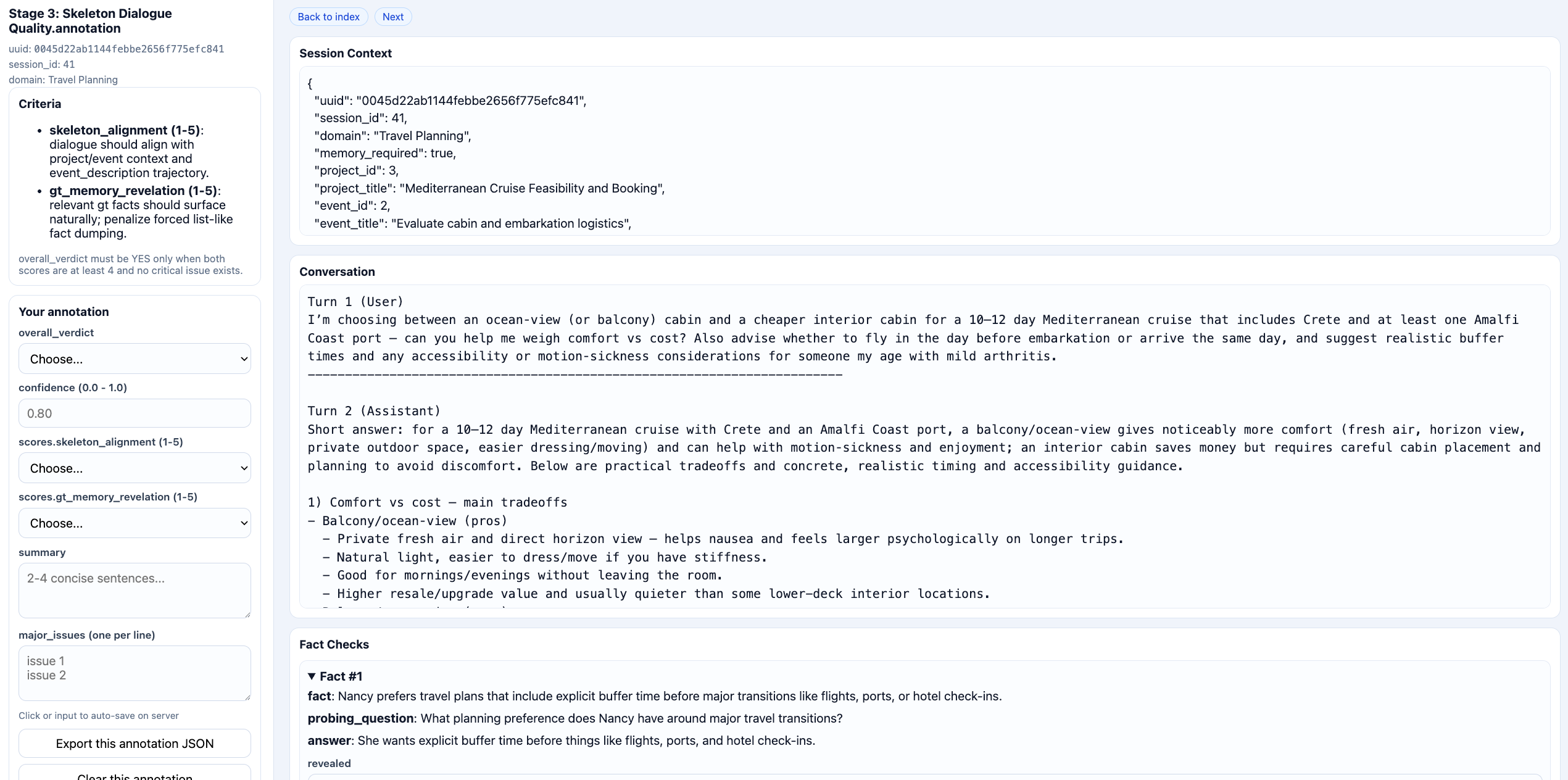}
    \caption{The user interface provided to the annotators for dialogue quality.}
    \label{fig:dialogue-annotation-ui}
\end{figure*}

The annotation interface provided to the human expert is shown in \Cref{fig:dialogue-annotation-ui}.

\section{Evaluation Protocol Details}
\label{sec:ap-rr-implementation}

\subsection{LLM-as-a-Judge for $\mathbb{I}(r \in \mathcal{M}_t)$}

To determine whether reference memory $r$ is present in memory bank 
$\mathcal{M}_t$, we use a two-step procedure. First, we retrieve the 
top-10 entries from $\mathcal{M}_t$ by cosine similarity using $r$ as 
the query (embedding model: \texttt{all-MiniLM-L6-v2}). Second, we 
pass these candidates to an LLM judge (\texttt{gpt-5-nano}) with the 
following prompt, which performs a binary classification on whether 
the core meaning of $r$ is preserved.

\begin{promptbox}[$\mathbb{I}(r \in \mathcal{M}_t)$ Judge Prompt]
\small
You are checking whether a specific piece of information is contained 
in a given text.

Fact to find: ``\{fact\}''

Text to search: \{retrieved\_entries\}

Does the text above contain the core meaning of this fact?
Answer YES if the fact is clearly expressed (even if worded differently).
Answer NO if the fact is absent or cannot be inferred from the text.

Reply with only YES or NO.
\end{promptbox}

\subsection{Checkpoint Sampling Approximation}

Computing $\mathbb{I}(r \in \mathcal{M}_t)$ at every session across 
all reference memories is prohibitively expensive: with up to 100 
sessions per user and ~50 reference memories per user, a full 
evaluation would require thousands of LLM calls per user per memory 
system. We therefore approximate the inner summation in \Cref{eq:rr}
using uniform checkpoint sampling.

Specifically, for each reference memory $r$, let 
$S(r) = [t_\text{start}(r),\, t_\text{start}(r){+}1,\, \ldots,\, 
T_\text{target}(r)]$ be the full set of evaluation sessions. 
We sample $K{=}20$ checkpoints $\hat{S}(r) \subset S(r)$ by selecting 
evenly spaced indices:
\begin{equation}
\hat{s}_i = S(r)\!\left[\,\text{round}\!\left(
  \frac{i \cdot (|S(r)|-1)}{K-1}
\right)\right], \quad i = 0, 1, \ldots, K{-}1,
\end{equation}
which guarantees that both the first session $t_\text{start}(r)$ and 
the last session $T_\text{target}(r)$ are always included. 
Including the endpoints is important because the first session 
verifies that the fact was stored at all, and the last session 
verifies that it survived until the end of its required lifespan.

The full inner sum is then approximated by reweighting the sampled 
scores:
\begin{equation}
\sum_{t=t_\text{start}(r)}^{T_\text{target}(r)} \mathbb{I}(r \in \mathcal{M}_t)
\;\approx\;
\frac{|S(r)|}{K} \sum_{t \in \hat{S}(r)} \mathbb{I}(r \in \mathcal{M}_t),
\end{equation}
where $\frac{|S(r)|}{K}$ is the rescaling factor under the 
assumption that each sampled checkpoint is representative of 
equally-sized intervals of the full evaluation window.
Substituting into Equation~(1) yields the approximated RR used in 
all experiments.

\section{Implementation Details}

\subsection{LLM Decoding}
To ensure reproducible and deterministic evaluation, we set 
temperature to 0 for all components: the memory systems under 
evaluation, the personalization methods, and the LLM-based 
data generation pipeline. For \texttt{Qwen3-14B}, inference is 
served via vLLM on A6000 48GB GPU. For proprietary models, we use their api services.

\subsection{Personalization Method}
\label{sec:ap-personalization-methods}

All three inference-based methods share a common interface: at each 
session, an LLM predicts \texttt{memory\_required} $\in \{\text{true}, 
\text{false}\}$, and sessions predicted as transient have their memory 
operations skipped. The methods differ only in what historical context 
is provided to the LLM.

\paragraph{Greedy.}
At each session, the LLM receives only the current session's dialogue 
(truncated to \texttt{max\_chars} characters) and outputs a binary 
prediction with no access to prior sessions.

\begin{promptbox}[Greedy Prediction Prompt]
\small
You are a memory policy agent. Decide whether this session should be 
stored in long-term memory.

\texttt{\#\# Current Session Dialogue}\\
\{current\_dialogue\}

\textbf{memory\_required = true}: long-horizon session — part of an 
ongoing project or recurring goal.\\
\textbf{memory\_required = false}: transient session — standalone 
and self-contained.

Respond ONLY with: \texttt{\{"memory\_required": <true|false>\}}
\end{promptbox}

\paragraph{Context-aware.}
Before prediction, the current session is summarized in 1--2 sentences 
and appended to a running summary buffer. At each session, the LLM 
receives the current dialogue alongside the most recent $K$ session 
summaries as historical context.

\begin{promptbox}[Context-aware Prediction Prompt]
\small
You are a memory policy agent. Decide whether this session should be 
stored in long-term memory.

\texttt{\#\# Recent Session Summaries (Past $K$ Sessions)}\\
\{history\_summaries\}

\texttt{\#\# Current Session Dialogue}\\
\{current\_dialogue\}

Respond ONLY with: \texttt{\{"memory\_required": <true|false>\}}
\end{promptbox}

\paragraph{Structure-aware.}
Each session is first parsed into a lightweight record 
$\{\texttt{purpose},\, \texttt{summary},\, \texttt{topic}\}$ via an 
LLM extraction call. Every $K$ sessions, these records are passed to 
a second LLM call that updates the structural note — clustering sessions 
into inferred projects or marking them as isolated. The note persists 
across windows (never reset), enabling retroactive reassignment of 
previously isolated sessions when new evidence connects them. 
Sessions not yet assigned by any completed window default to 
\texttt{memory\_required = true} conservatively.

\begin{promptbox}[Session Record Extraction Prompt]
\small
Analyze this conversation session and extract a structured summary.

\texttt{\#\# Session Dialogue}\\
\{dialogue\}

Respond ONLY with:
\texttt{\{"purpose": "...", "summary": "...", "topic": "..."\}}
\end{promptbox}

\begin{promptbox}[Structural Note Update Prompt]
\small
You are managing a structural note that tracks a user's ongoing 
projects across AI assistant sessions.

\texttt{\#\# Current Structural Note}\\
\{current\_note\}

\texttt{\#\# New Session Records (session\_id \{start\}$\sim$\{end\})}\\
\{session\_records\}

Rules:\\
1. Group sessions belonging to the same ongoing project.\\
2. Self-contained one-off sessions go in \texttt{isolated\_sessions}.\\
3. Previously isolated sessions MAY be reassigned to a project if 
   new evidence connects them.\\
4. Every session\_id must appear in exactly one place.

Respond ONLY with the updated note:
\begin{verbatim}
{
  "projects": [
    {"project_id": "P1", "label": "...",
     "core_topic": "...", "session_ids": [...],
     "status": "ongoing"|"completed"}
  ],
  "isolated_sessions": [...]
}
\end{verbatim}
\end{promptbox}

\section{Further Analyses}

\subsection{Sensitivity Analysis of $K$ in Estimating Memory Retention Rate}

We conduct a sensitivity analysis by varying the number of checkpoints $K \in \{10,20,40\}$  and comparing against the full (unsampled) computation on \proposeddata. As shown in \Cref{tab:checkpoint_sensitivity}, the absolute difference in RR between $K$=20 and Full is negligible, indicating that our approximation incurs almost no loss in accuracy while substantially reducing computational cost. As expected, reducing the checkpoint count further ($K$=10) introduces somewhat more instability, while increasing it ($K$=40) yields results comparable to our default ($K$=20). This suggests that $K$=20 offers a reasonable balance between computational cost and estimation accuracy, noting that practitioners with different resource constraints may adjust $K$ accordingly based on this trade-off.

\begin{table}[h]
\centering
\caption{Sensitivity analysis of retention rate (RR) across different numbers of evaluation checkpoints ($K$). The default setting ($K=20$) is highlighted in bold.}
\label{tab:checkpoint_sensitivity}
\resizebox{0.7\columnwidth}{!}{
\begin{tabular}{llcccc}
\toprule
\textbf{Backbone} & \textbf{Setting} & \textbf{$K=10$} & \textbf{$K=20$} & \textbf{$K=40$} & \textbf{Full} \\
\midrule
\multirow{2}{*}{Mem0}
    & Base   & 0.5343 & \textbf{0.5233} & 0.5272 & 0.5258 \\
    & Oracle & 0.6257 & \textbf{0.6183} & 0.6265 & 0.6220 \\
\midrule
\multirow{2}{*}{Memory-R1}
    & Base   & 0.5733 & \textbf{0.5600} & 0.5651 & 0.5594 \\
    & Oracle & 0.6027 & \textbf{0.6063} & 0.6056 & 0.6033 \\
\bottomrule
\end{tabular}
}
\end{table}

\clearpage
\newpage

\end{document}